\documentclass[journal]{IEEEtran}
\usepackage[utf8]{inputenc}
\usepackage{amsmath,amssymb}
\usepackage{graphicx}
\usepackage{geometry}
\usepackage{hyperref}
\usepackage{stfloats}
\usepackage{float}  % for [H] placement
\usepackage{booktabs}
\usepackage{gensymb}
\usepackage{textcomp} % Load this package in the preamble
\usepackage{array}
\usepackage{longtable}
\usepackage{tabularx}
\usepackage[table]{xcolor}
\definecolor{rowgray}{gray}{0.96}  % very light gray
\definecolor{headerblue}{RGB}{230,243,255}  % very light blue

\title{Sparse-View 3D Reconstruction: Recent Advances and Open Challenges}
\author{
    Tanveer Younis \\
    Zhanglin Cheng \\
    \textit{Shenzhen Institute of Advanced Technology, Chinese Academy of Sciences} \\
    \textit{Shenzhen VisuCA Key Lab, Shenzhen, China} \\
    \texttt{\{younis, zl.cheng\}@siat.ac.cn}
}

\begin{document}

\maketitle

\begin{abstract}

Sparse-view 3D reconstruction is essential for applications in which dense image acquisition is impractical, such as robotics, augmented/virtual reality (AR/VR), and autonomous systems. In these settings, minimal image overlap prevents reliable correspondence matching, causing traditional methods, such as structure-from-motion (SfM) and multiview stereo (MVS), to fail. This survey reviews the latest advances in neural implicit models (e.g., NeRF and its regularized versions), explicit point-cloud-based approaches (e.g., 3D Gaussian Splatting), and hybrid frameworks that leverage priors from diffusion and vision foundation models (VFMs).We analyze how geometric regularization, explicit shape modeling, and generative inference are used to mitigate artifacts such as floaters and pose ambiguities in sparse-view settings. Comparative results on standard benchmarks reveal key trade-offs between the reconstruction accuracy, efficiency, and generalization. Unlike previous reviews, our survey provides a unified perspective on geometry-based, neural implicit, and generative (diffusion-based) methods. We highlight the persistent challenges in domain generalization and pose-free reconstruction and outline future directions for developing 3D-native generative priors and achieving real-time, unconstrained sparse-view reconstruction.

\end{abstract}

\section{Introduction}
\label{sec:introduction}
Reconstructing three-dimensional (3D) scenes from two-dimensional (2D) images has been a central challenge in computer vision for decades. Early approaches, such as structure-from-motion (SfM)\cite{schoenberger2016sfm} and Multiview Stereo (MVS)\cite{schoenberger2016mvs}, typically depended on dense, highly overlapping sets of images to achieve reliable results. However, in many real-world scenarios, such as robotics, augmented reality (AR), virtual reality (VR), autonomous navigation, and digital content creation, collecting such dense image datasets is often difficult or costly. Consequently, research has increasingly focused on sparse-view 3D reconstruction, where the goal is to produce accurate and detailed 3D models using only a small number of partially overlapping images.

\begin{figure}[htbp]
    \centering
    \includegraphics[width=\columnwidth]{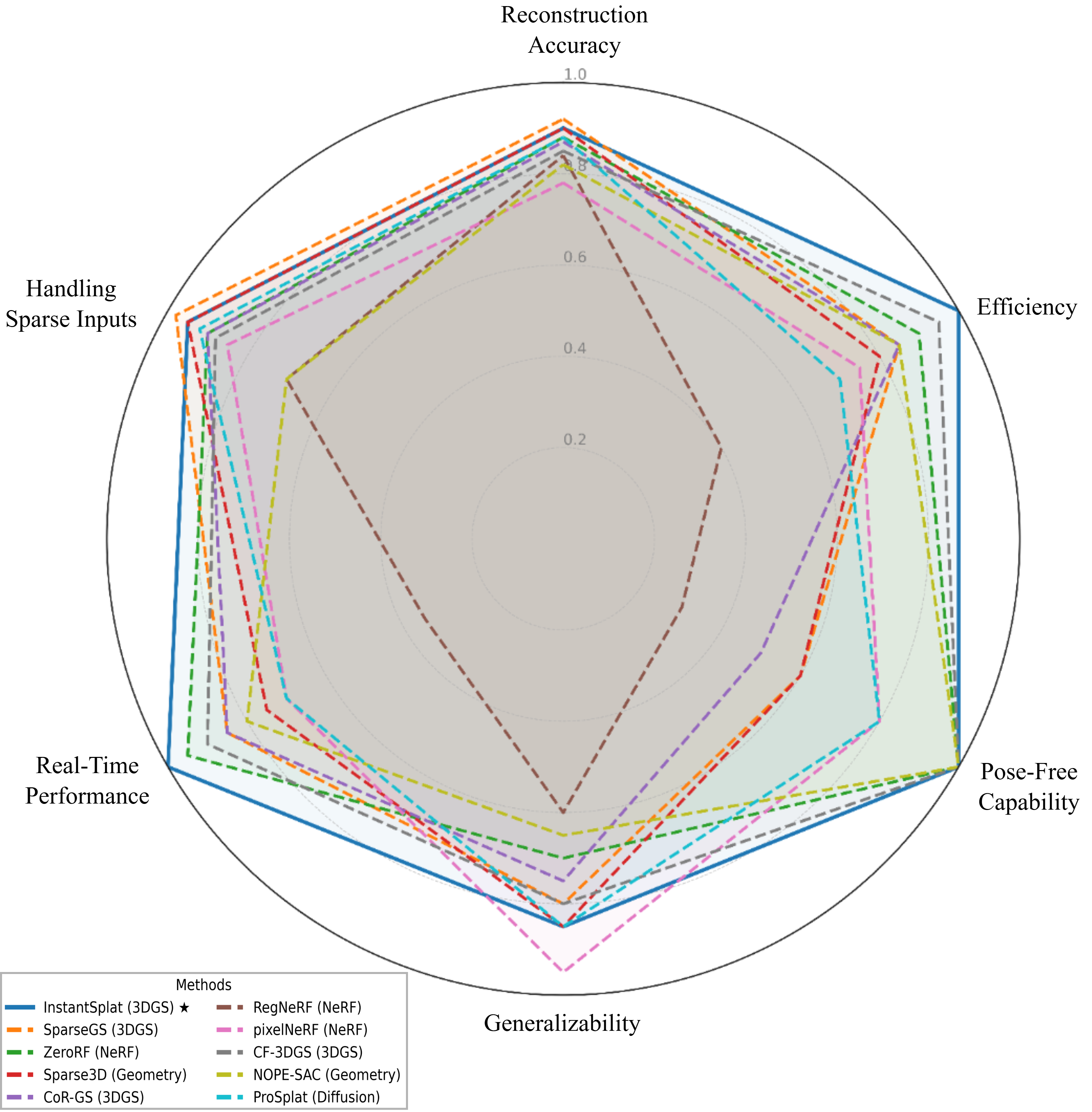}
    \caption{Comparative performance of leading sparse-view 3D reconstruction methods across six normalized metrics: Handling Sparse Inputs, Pose-Free Capability, Real-Time Performance, Efficiency, Generalizability, and Reconstruction Accuracy (all scores normalized to [0–1] scale, where 1.0 denotes highest performance).}
    \label{fig:radar_chart}
\end{figure}

While existing surveys have addressed broader aspects of 3D reconstruction\cite{RABBY2023BeyondPixelsAC,Farshian2023DeepLearningBased3S,Wu2024Multiview3R,wu2024recent} or focused on specific techniques like 3D Gaussian Splatting for sparse views\cite{Liu2025ARO}. To the best of our knowledge, no previous study has systematically analyzed the convergence of geometry-based, neural implicit, and generative (diffusion-based) approaches in sparse-view 3D reconstruction. Our survey addresses this gap by providing a unified framework and comparative evaluation of all leading classes of methods.

Sparse-view 3D reconstruction is inherently ambiguous because of limited input, leading to artifacts such as floaters, blurred textures, background collapse, and pose estimation ambiguity\cite{chanPointCloudDensification2024}. This persistent 'chicken-and-egg' problem, which becomes particularly severe with limited input views, has shifted the research focus towards deep learning methods that can jointly optimize or bypass explicit pose estimation.

\begin{figure*}[h]  % [h] for here; can also use [t] or [b]
    \centering
    \includegraphics[width=\linewidth]{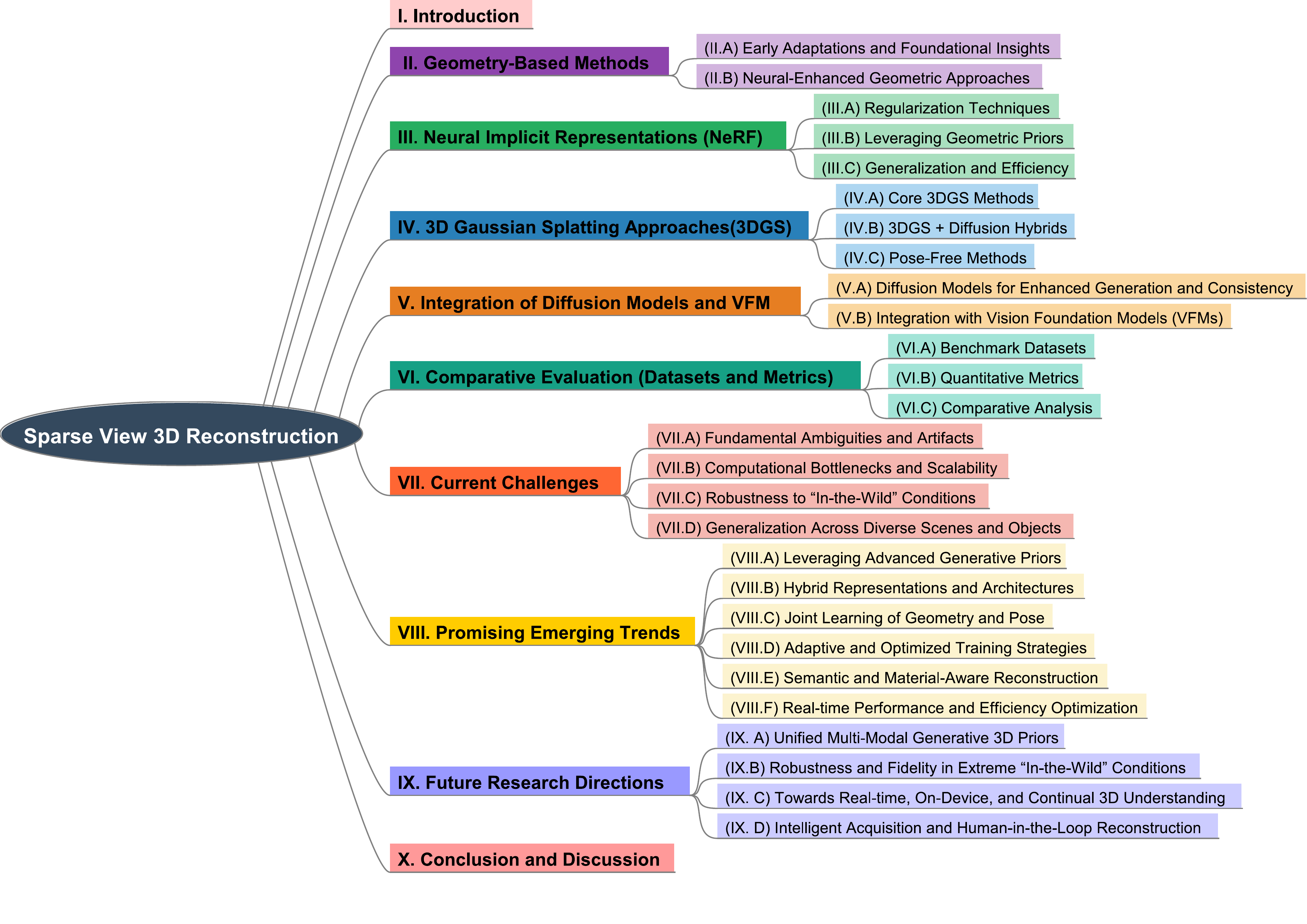}
    \caption{Structure of this survey: major topics and subtopics covered in sparse-view 3D reconstruction.}
    \label{fig:structure}
\end{figure*}

Deep learning-based methods have recently led to significant advances in both reconstruction quality and robustness. Implicit neural representations, such as Neural Radiance Fields (NeRFs)\cite{mildenhall2020nerf}, and explicit representations, such as 3D Gaussian Splatting (3DGS)\cite{kerbl2023threedgs}, have driven much of this progress. NeRF, in particular, has had a major impact on sparse-view reconstruction by encoding scenes as continuous volumetric functions, enabling the synthesis of realistic novel views from only a handful of images\cite{wangSparseNeRFDistillingDepth2023, shiZeroRFFastSparse2023}. While early NeRF variants struggled with computational inefficiency and overfitting, newer methods have incorporated depth priors\cite{liDNGaussianOptimizingSparseView2024}, geometric regularization\cite{younesSparseCraftFewShotNeural2024, xiangPointGSPointAttentionAware2025}, and semantic consistency. Collectively, these advances enable significantly improved results with fewer input views than those of prior studies.

\begin{figure*}[t]  % use [t] or [b] for top or bottom placement
    \centering
    \includegraphics[width=\textwidth]{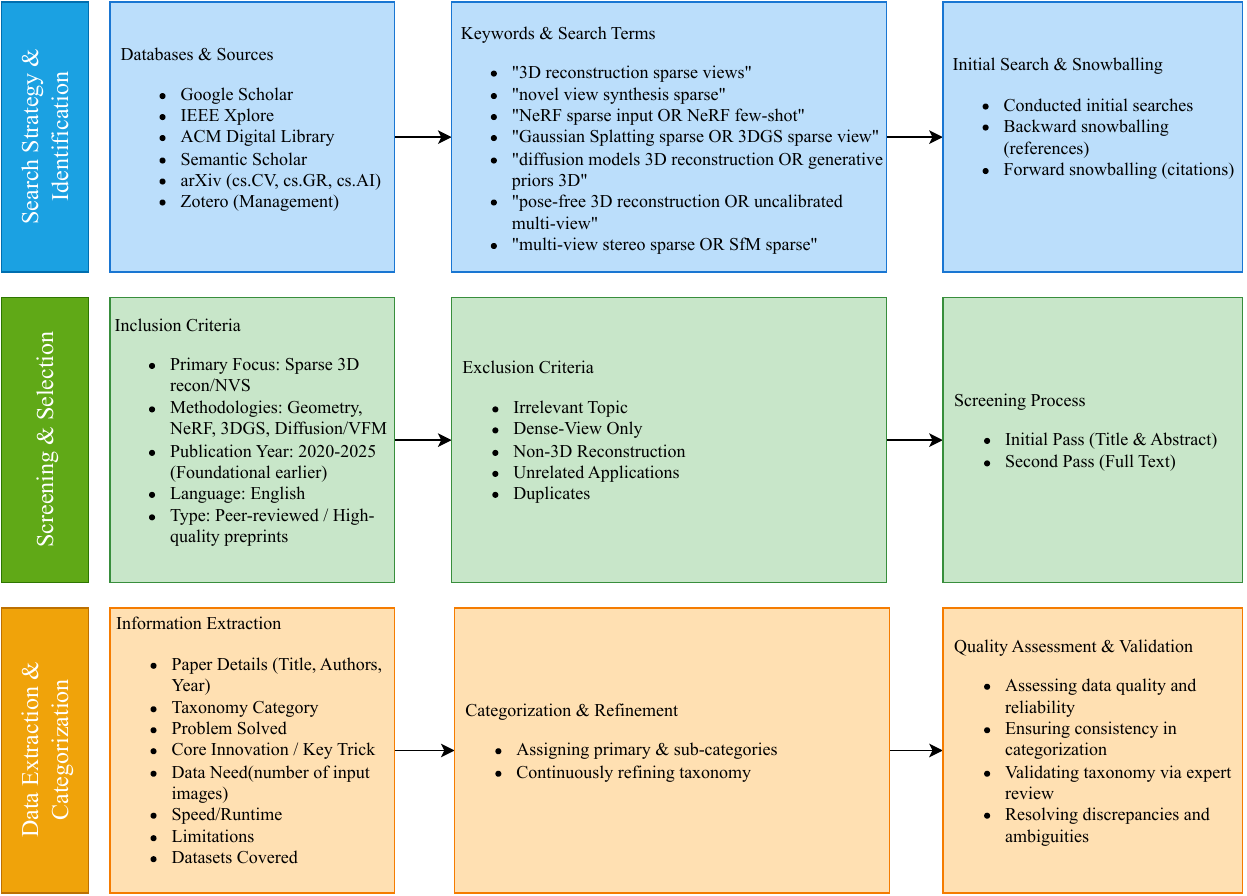}
    \caption{Methodological Review Protocol outlining the systematic process of literature identification, screening, data extraction, categorization, and quality assessment used in this sparse-view 3D reconstruction survey.}
    \label{fig:search}
\end{figure*}

\begin{figure*}[t]  % use [t] or [b] for top or bottom placement
    \centering
    \includegraphics[width=\textwidth]{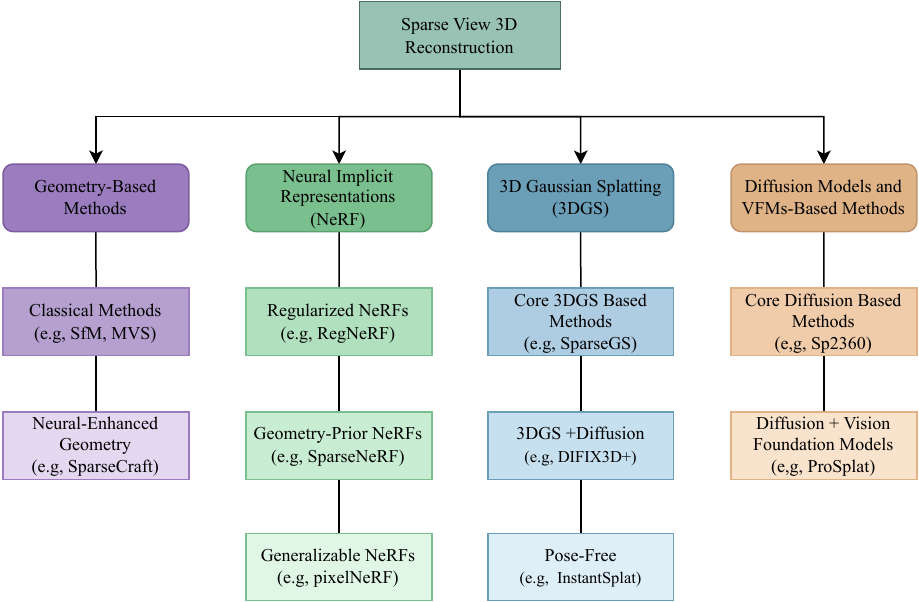}
    \caption{Taxonomy of sparse view 3D reconstruction methods by core categories.}
    \label{fig:taxonomy}
\end{figure*}

Recent advances in explicit representations, especially 3DGS\cite{kerbl2023threedgs}, have resulted in substantial gains in computational efficiency and real-time rendering. By modeling scenes with Gaussian primitives, 3DGS allows for fast rasterization into images\cite{fanInstantSplatUnboundedSparseview,fuCOLMAPFree3DGaussian}. New methods use depth-informed pruning and co-regularization\cite{zhangCoRGSSparseView3D2024,zhuFSGSRealTimeFewshot2024} to reduce overfitting and limit artifacts, particularly when input images are sparse. InstantSplat\cite{fanInstantSplatUnboundedSparseview} demonstrated that high-quality reconstructions can be completed in a few seconds. This demonstrates significant improvements in both the speed and robustness to errors in the camera pose.

Recent studies have shown that diffusion-based generative models reduce ambiguity in sparse-view reconstruction by predicting the likely shapes and textures. Diffusion models\cite{ho2020ddpm}, trained on extensive datasets, provide strong priors that improve the realism and consistency of both images and 3D outputs\cite{gaoCAT3DCreateAnything2024}. Researchers have combined these generative models with NeRF and 3DGS in hybrid systems. This blending of explicit and implicit representations enables a better balance between quality, efficiency, and usability\cite{barthel2024gaussian}. Camera pose estimation is a central challenge in sparse-view 3D reconstruction, motivating the development of pose-free methods that directly recover geometry from uncalibrated images. Recent approaches such as InstantSplat\cite{fanInstantSplatUnboundedSparseview}, COLMAP-Free 3D Gaussian Splatting\cite{fuCOLMAPFree3DGaussian}, and MV-DUSt3R+\cite{tangMVDUSt3RSingleStageScene} exemplify this trend. These methods enable robust 3D reconstruction even in difficult image-capture scenarios. 

\begin{figure}[htbp]
    \centering
    \includegraphics[width=\columnwidth]{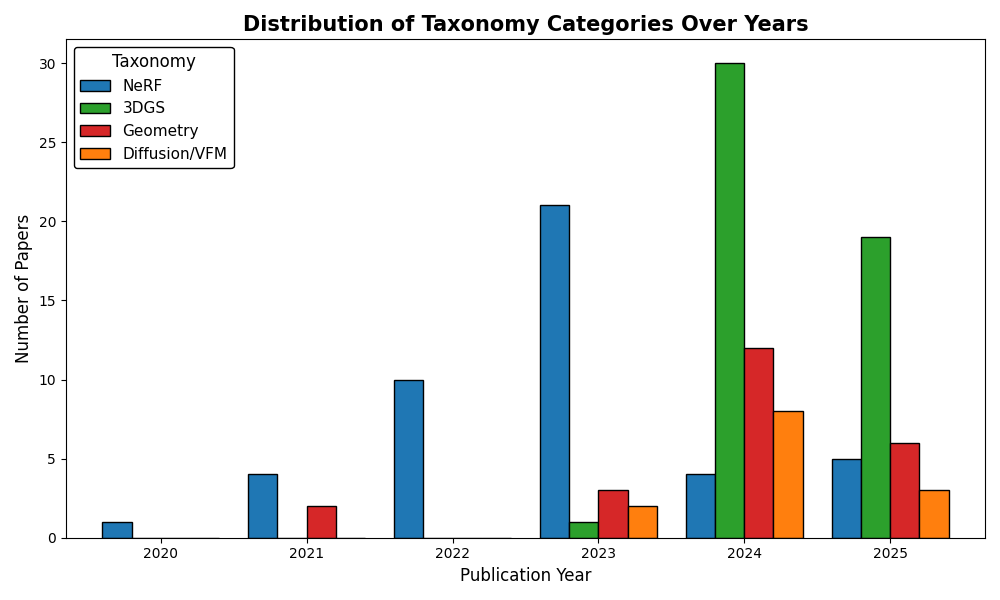}
    \caption{Distribution of sparse-view 3D reconstruction papers (2020–2025) by taxonomy category. The exponential growth in 3DGS and Diffusion/VFM papers post-2022 reflects their breakthrough efficiency (3DGS) and ability to synthesize missing information (Diffusion/VFM), which directly addresses the limitations of earlier NeRF variants (computational cost, overfitting with sparse inputs).}
    \label{fig:timeline}
\end{figure}

This survey reviews recent advances in sparse-view 3D reconstruction, focusing on core technical challenges and how new methods—spanning geometric priors, diffusion models, and improved representations—have advanced the field. To contextualize this progression, figure~\ref{fig:timeline} illustrates the development and relative prominence of these categories over time. We also summarize the key performance benchmarks and discuss persistent problems. Finally, we outline promising research directions that may help address these issues. The overall structure of this study is illustrated in Figure \ref{fig:structure}. The main contributions of this review are as follows.
\begin{itemize}
    \item \textbf{Systematic Categorization}: We organize recent sparse-view 3D reconstruction methods into geometry-based, neural implicit (NeRF), 3D Gaussian Splatting (3DGS), and hybrid classes, clearly outlining core mechanisms and limitations.
    \item \textbf{In-depth Analysis of 3DGS Methods}: We present the most extensive and up-to-date review of 3D Gaussian Splatting techniques, including core, diffusion-integrated, and pose-free variants, with a focus on their effectiveness in sparse-view settings.
    \item \textbf{Integration of Generative Models}: We analyze how diffusion models and vision foundation models (CLIP, SAM, DINO) are being leveraged to inject strong priors, enforce view consistency, and hallucinate plausible geometry from limited data.
    \item \textbf{Cross-paradigm Comparison}: We provide critical comparisons across paradigms (SfM, NeRF, 3DGS, diffusion), evaluating their trade-offs in accuracy, efficiency, generalizability, and real-world applicability under sparse constraints.
    \item \textbf{Identification of Research Gaps}: We outline unresolved challenges such as domain generalization, pose-free reconstruction, and efficient learning from minimal supervision, paving the way for future research directions.
\end{itemize}
The complete review process, including the search strategy, literature selection, screening, data extraction, categorization, and quality assessment, is shown in figure~\ref{fig:search}.

\section{Geometry-Based Methods}
\label{sec:geometry_methods}

Traditional 3D reconstruction pipelines, mainly based on  SfM \cite{schoenberger2016sfm} and MVS\cite{schoenberger2016mvs}, have long formed the backbone of visual 3D scene understanding. However, these classical methods often struggle with sparse views, where there may be too few correspondences, making accurate reconstruction much harder~\cite{tanNOPESACNeuralOnePlane2023,jin3DFIRESFewImage2024,weiSemanticallyAwareMultiView2024,saxena3DReconstructionSparse2007}. In this section, we review the progression from early attempts to recent neural-enhanced geometric approaches that aim to overcome these challenges.

\subsection{Early Adaptations and Foundational Insights}

Early methods sought to improve multiview triangulation in sparse-view settings by incorporating extra cues. For example, \cite{saxena3DReconstructionSparse2007} proposed a Markov Random Field framework that combines monocular cues with multiview triangulation for the joint inference of 3D position and orientation. By modeling geometric relationships, such as collinearity and coplanarity, and applying occlusion constraints, this approach marked a significant improvement over methods that relied solely on dense correspondences.

Chen et al. \cite{chenSingleSparseView2011} further pushed the boundaries of single- and sparse-view 3D reconstruction with a framework based on Gaussian Process Latent Variable Models (GPLVM)~\cite{chenSingleSparseView2011}. Their method learns shape priors from a collection of training examples so that during inference, a new silhouette can be matched to the learned shape space. This allows for plausible 3D shape recovery, even with a very limited input. By regularizing the reconstruction problem and modeling both variability and uncertainty in shapes, this technique is particularly useful for object categories with a consistent structure and degree of geometric flexibility. Schönberger et al.~\cite{schoenberger2016mvs} introduced pixel-wise view selection for unstructured multi-view stereo, a method designed to improve the quality of dense 3D reconstruction through pixel-wise selection of the input images. This approach selects the most reliable source images for each pixel, thereby improving the dense reconstruction. While designed for unstructured dense sets, its strategies for managing view redundancy and optimal view selection are also relevant to sparse-view MVS

\subsection{Neural-Enhanced Geometric Approaches}

Structure-from-Motion ~\cite{schoenberger2016sfm} offers a thorough review and critical assessment of SfM techniques. Although not dedicated to sparse-view settings, this study is a foundational reference for the principles behind SfM, which underpins pose estimation in several 3D reconstruction pipelines. The discussion of the strengths of SfM and its sensitivity to feature correspondences helps to explain the difficulties faced by traditional methods under sparse conditions.

Recent geometry-based methods have addressed these limitations by incorporating neural networks, enabling the overcoming of issues such as poor feature correspondence, dynamic scene content, and incomplete geometry. Jin et al.~\cite{Jin2021PlanarSR} proposed a learning-based approach for reconstructing structured planar surfaces from two unposed RGB images. Their method is tailored for cases in which conventional multi-view geometry fails, such as when the baselines are wide or the observations sparse. The system jointly estimates the planar hypotheses, cross-view correspondences, and relative 6-DoF camera poses. A deep network predicts planar segments and cross-view embeddings, which are then refined using a two-stage discrete-continuous optimization. This unified pipeline fuses geometric and pose estimates to yield a 3D reconstruction, demonstrating a strong performance in indoor scenes with minimal viewpoint overlap.

NOPE-SAC (Neural One-Plane RANSAC) from Tan et al.~\cite{tanNOPESACNeuralOnePlane2023} made significant strides in two-view 3D reconstruction by using neural networks to learn pose hypotheses from limited plane correspondences. This method addresses the challenges arising from severe viewpoint changes and low-texture scenes by employing an end-to-end RANSAC-like process that delivers reliable camera pose estimates and reconstruction accuracy. Notably, it outperforms classical approaches on standard benchmarks, such as Matterport3D \cite{chang2017matterport3d} and ScanNet. Mu et al.~\cite{muNeural3DReconstruction2023} presented a neural implicit framework that explicitly uses geometric priors, such as depth and surface normals, to improve sparse-view 3D surface reconstruction. Their method surpasses both classical MVS systems, such as COLMAP, and neural baselines, including PixelNeRF~\cite{yu2021pixelnerf}, excelling in terms of geometric detail and convergence speed.

\begin{table*}[h]
\centering
\scriptsize
\renewcommand{\arraystretch}{1.2}
\rowcolors{2}{rowgray}{white}
\setlength{\tabcolsep}{6pt}
\resizebox{\textwidth}{!}{%
\begin{tabular}{l|c|c|c|c|c}
\toprule
\textbf{Method} & \textbf{Year} & \textbf{Input Views} & \textbf{Pose Needed} & \textbf{Representation Type} & \textbf{Runtime} \\
\midrule
Stereo Radiance Fields (SRF) \cite{chibaneStereoRadianceFields2021} & 2021 & Sparse (10) & Yes & Neural Radiance Field (MVS-inspired) & Fast (fine-tune in min) \\
NOPE-SAC  \cite{tanNOPESACNeuralOnePlane2023} & 2023 & 2 & No & Neural One-Plane RANSAC & Fast (pose estimation) \\
Neural 3D reconstruction from sparse views  \cite{muNeural3DReconstruction2023} & 2023 & Sparse (3-7) & Yes & Neural Implicit (Geometric Priors) & Fast (faster convergence than baselines) \\
A Semantically Aware Multi-View 3D Reconstruction  \cite{weiSemanticallyAwareMultiView2024} & 2024 & Multi-view & Yes & SfM/SGM + Semantic Labels & Moderate (improved accuracy, not speed) \\
SparseCraft \cite{younesSparseCraftFewShotNeural2024} & 2024 & Few-shot (3-9) & Yes & Implicit SDF (Stereopsis-guided) & Fast \\
3DFIRES  \cite{jin3DFIRESFewImage2024} & 2024 & Few-shot (1-3) & Yes & DRDF + Transformer (Feature Fusion) & Moderate (due to transformer backbone) \\
Dust to Tower  \cite{caiDustTowerCoarsetoFine2024} & 2024 & Sparse, uncalibrated & No & Coarse-to-fine framework (CCM, CADA, WIGI) & Fast (reconstruction in seconds) \\
GS4  \cite{jiangGS4GeneralizableSparse2025} & 2025 & RGB-D video stream & Yes & Sparse Splatting Semantic SLAM & Real-time (rendering) \\
Neural Surface Reconstruction \cite{zhouNeuralSurfaceReconstruction2024} & 2024 & Sparse & Yes & Epipolar information + Monocular Depth Priors & Efficient (training/inference) \\
\bottomrule
\end{tabular}%
}
\caption{Comparison of Geometry-Based Methods for Sparse-View 3D Reconstruction. Runtimes are as reported in original works and refer to training or inference as indicated.}
\label{tab:geometry_methods_comparison}
\end{table*}

SparseCraft by Younes et al. ~\cite{younesSparseCraftFewShotNeural2024} adopts a different approach, employing stereopsis-guided geometric linearization within an implicit Signed Distance Function (SDF) framework and using normals and colors derived from MVS for regularization. This results in state-of-the-art performance on few-shot reconstruction tasks, offering both speed and robustness from sparse inputs.

Stereo Radiance Fields (SRF) from Chibane et al.~\cite{chibaneStereoRadianceFields2021} generalize neural view synthesis by implicitly learning photoconstistent scene structures from sparse input views. The SRF model can infer scene geometry and generate high-quality colored meshes without requiring extensive retraining, providing a strong example of recent progress in this area.

Wei et al.~\cite{weiSemanticallyAwareMultiView2024} introduced a semantically aware multiview 3D reconstruction method that integrates semantic labels into SfM and Semi-Global Matching (SGM) \cite{hirschmuller2005sgm} pipelines. By applying semantic-based filtering and segmentation, their approach significantly improved the reconstruction accuracy, particularly in dynamic urban environments, which is an important consideration for autonomous driving and related applications.

Jin et al.~\cite{jin3DFIRESFewImage2024} tackled the challenge of hidden surface reconstruction with 3DFIRES, a system that fuses multi-view information at the feature level to reason about occluded regions. Leveraging a Directed Ray Distance Function (DRDF)\cite{kulkarni2022directed} and a transformer-based architecture\cite{ranftl2021vision, rombach2021geometry}, 3DFIRES can achieve comprehensive and accurate reconstructions from extremely sparse image sets, consistently outperforming previous methods. Advances in neural-enhanced geometry-based methods have shown a progression from basic augmentations to the integration of learned priors, semantic understanding, and advanced geometric reasoning. Collectively, these methods expand the practical applicability and robustness of 3D reconstruction in sparse-view scenarios.

Extending these neural-enhanced approaches, Zhou et al.~\cite{zhouNeuralSurfaceReconstruction2024} introduced Neural Surface Reconstruction from Sparse Views Using Epipolar Geometry (EpiS)\cite{hartley2004multiple}. This method incorporates epipolar information to improve surface reconstruction from sparse inputs by aggregating coarse information from cost volumes into epipolar features across multiple views to generate detailed signed distance function (SDF)-aware features. EpiS uses pretrained monocular depth models with global triplet and local gradient losses for regularization, outperforming both state-of-the-art neural implicit and classical MVS methods, particularly in sparse and generalizable settings. Zhang et al.~\cite{Hong20243DCH} addressed the problem of reconstructing 3D clothed humans from sparse multi-view images, a particularly challenging task due to non-rigid deformations and complex clothing. Their method offers practical solutions for dynamic and deformable objects, areas in which both traditional and neural approaches often struggle.

Zhu et al.\cite{Ye2024PVPReconPV} introduced PVP-Recon, which uses progressive view planning and warping consistency to select optimal viewpoints and incrementally improve sparse-view surface reconstruction. This method illustrates how active geometry-driven view selection can address the challenges of sparse-view reconstruction. Table~\ref{tab:geometry_methods_comparison} compares representative geometry-based methods by year, input requirements, pose dependency, representation type, and runtime.

\section{Neural Implicit Representations (NeRF and Variants)}
\label{sec:nerf_methods}

While the original NeRF\cite{mildenhall2020nerf} enabled highly photorealistic rendering from dense input views, its performance degraded significantly under sparse-view conditions owing to overfitting and geometric inaccuracies. This section reviews recent advances in NeRF and its variants. We focused on methods developed to address the challenges posed by sparse views. The key strategies include regularization, the use of geometric and learned priors, and improvements in the generalization and computational efficiency.

\subsection{Regularization Techniques}

Regularization methods are essential for stabilizing NeRF training, particularly when the input views are sparse. RegNeRF~\cite{niemeyerRegNeRFRegularizingNeural2022} addresses this issue by introducing both geometric and color regularizations. It uses a patch-based geometry regularizer and a pretrained normalizing flow model\cite{dinh2017realnvp} for color prediction. An annealing strategy for ray sampling was also employed. Together, these steps reduce floating artifacts and improve geometric consistency.

FlipNeRF ~\cite{seoFlipNeRFFlippedReflection2023} builds on this by using flipped reflection rays to generate richer training data. It introduces an uncertainty-aware emptiness and bottleneck feature consistency losses. These changes significantly enhance geometric fidelity and help reduce rendering artifacts. Liu et al.~\cite{Lu2024FastSV} propose a fast update mechanism for NeRF, aimed at object reconfiguration from sparse views. This method is particularly useful for dynamic scenes or situations in which the content of the scene changes. This improves the robustness and adaptability of NeRFs in non-static sparse-view environments.

AS-NeRF, proposed by Zhang et al.~\cite{Tang2024ASNeRFLA}, learns auxiliary sampling strategies for generalizable novel view synthesis from sparse views. By improving the sampling efficiency and coverage, AS-NeRF delivers better reconstruction quality and generalization without requiring dense input. SC-NeRF~\cite{Song2023SCNeRFSN} introduced a self-correcting mechanism for NeRF models trained using sparse views. The method identifies and fixes inconsistencies or artifacts that result from limited data, improving both robustness and scene fidelity. Sparse-DeRF, developed by Lee et al.~\cite{Lee2024SparseDeRFDN}, addresses the problem of blurred renderings in NeRFs trained on sparse data. It incorporates a deblurring mechanism directly into the NeRF framework, resulting in sharper and more photorealistic novel views, even with limited input images.

\subsection{Leveraging Geometric Priors}

Integrating external geometric priors is a common and effective method for guiding NeRF optimization when the input views are sparse. Roessle et al.~\cite{roessleDenseDepthPriors2022} introduced dense depth priors obtained from SfM, along with pixel-level uncertainty estimation. These priors constrain ray termination in the volume rendering of the NeRF, significantly reducing geometric errors and improving image quality, even with very few input views.

DS-NeRF~\cite{dengDepthsupervisedNeRFFewer2022} relies on depth supervision using sparse SfM point clouds. It applies a probabilistic loss based on KL divergence, which accelerates convergence and improves depth accuracy without requiring dense ground truth data. SparseNeRF~\cite{wangSparseNeRFDistillingDepth2023} distills depth-ranking priors from monocular depth estimators. By enforcing local depth consistency and smoothness, this method achieves accurate reconstruction, even in severely sparse-view scenarios.

Mostegel et al.~\cite{Zhang2023SparseSatNeRFDD} developed SparseSat-NeRF, which uses dense depth supervision for reconstructing scenes from sparse satellite images. This approach leverages depth priors to improve the results of aerial imaging, where wide baselines and sparse views are common, and a dedicated network converts sparse 3D points into dense depth maps and pixel-wise uncertainty estimations. These are then used in NeRF training via Gaussian Negative Log Likelihood (GNLL) depth loss and depth-guided sampling. This setup enables efficient novel view synthesis, particularly for room-scale scenes.

\subsection{Generalization and Efficiency}

Many NeRF variants have been designed for generalizable reconstruction and faster inference, moving beyond the per-scene optimization. While PixelNeRF  ~\cite{yu2021pixelnerf} enabled feed-forward generalization, methods like IBRNet ~\cite{wangIBRNetLearningMultiView2021} further improved efficiency through image-based rendering principles, and ZeroRF ~\cite{shiZeroRFFastSparse2023} eliminated lengthy pre-training for 360° scenes.

X-NeRF~\cite{zhuXNeRFExplicitNeural2022} focuses on sparse and nearly non-overlapping RGB-D views. It uses a generative CNN to complete sparse RGB-D tensors, allowing robust generalization across scenes and fast inference without additional optimization. 6Img-to-3D~\cite{gieruc6Imgto3DFewImageLargeScale2025} is highly efficient at reconstructing large-scale outdoor scenes from very sparse inputs. It uses transformer-based encoding and differentiable volume rendering to provide an accurate depth and novel views without explicit depth supervision. This makes it particularly useful in autonomous driving scenarios. NeRS~\cite{zhangNeRSNeuralReflectance} introduced a surface-based representation designed for sparse-view 3D object reconstruction from in-the-wild images. It explicitly models surfaces using neural reflectance functions, making the reconstruction robust against noisy camera poses and challenging lighting. NeRS consistently outperformed volumetric methods in real-world applications. 

Zhu et al.~\cite{zhuVanillaMLPNeural2024} build on regularization strategies by questioning whether the standard MLP in NeRF is enough for few-shot view synthesis. They introduced a multi-input MLP (mi-MLP), which feeds both the location and viewing direction into each layer. This simple change helps prevent overfitting without losing details. This method also models the color and volume density separately and adds two regularization terms to reduce artifacts, achieving state-of-the-art results across several benchmarks.

CMC~\cite{zhuCMCFewshotNovel2024} tackles overfitting in few-shot NeRFs with a different approach. It enforces depth-aware consistency across the input views by ensuring that the same spatial points are sampled repeatedly in different images. The method uses neural networks on layered (multiplane) representations and constrains both the color and depth to be consistent across the views. This led to improved synthesis quality compared with previous methods. CVT-xRF~\cite{zhongCVTxRFContrastiveInVoxel2024} focuses on enhancing 3D spatial consistency in NeRF under sparse input. It uses a voxel-based ray-sampling strategy and a Contrastive In-Voxel Transformer (CVT). The transformer infers ray point properties from the local voxel context, whereas voxel contrastive regularization enforces feature similarity between neighboring regions. This approach greatly improves consistency and removes artifacts.

Bao et al.~\cite{baoWhereHowMitigating2023} address the issue of "confusion" in sparse-input NeRFs, which can lead to overfitting and foggy surfaces. Their method addresses the questions of "where to sample" and "how to predict" before volume rendering. They introduced a deformable sampling strategy with mutual information loss to reduce the sample position confusion. A semi-supervised paradigm based on pose perturbation combined with pixel-patch correspondence loss addresses prediction confusion. This method achieves state-of-the-art performance without the need for pretrained models or computationally intensive warping. NeRF-OR~\cite{Gerats2024NeRFORNR} introduced a method for 3D scene reconstruction from sparse-view RGB-D videos, specifically tailored for operating room (OR) environments. This approach combines the time-of-flight sensor depth with dense depth estimates\cite{ke2024repurposing} from color images and uses surface normals derived from these depths for the supervision. NeRF-OR accurately reconstructs dynamic surgical scenes, captures fine geometric details, and trains significantly faster than previous methods. It also generalizes well to other sparse-view reconstruction benchmarks. DaRF~\cite{songDaRFBoostingRadiance2023} proposed a framework that combines NeRF with monocular depth estimation (MDE)\cite{ranftl2021vision} through online complementary training. This method imposes the MDE geometric priors on the NeRF at both the seen and unseen viewpoints. DaRF addresses ambiguities in MDE using patch-wise scale-shift fitting and geometry distillation, adapting the MDE network to align with NeRF geometry. This results in robust reconstructions from a few images and achieves state-of-the-art performance. 

DiViNeT~\cite{voraDiViNeT3DReconstruction2023} tackles 3D reconstruction from sparse and disparate views using learned neural templates as surface prior. The approach works in two stages: first, it learns templates as 3D Gaussian functions across scenes; then, it uses these templates as anchors to complete the surface geometry and recover details. DiViNeT is especially effective in cases where radiance ambiguity causes traditional methods to fail, producing high-quality reconstructions from as few as three images.

Li et al.~\cite{liRegularizingNeuralRadiance2023} introduce a regularization framework for NeRF that leverages both geometry and appearance cues, guided by depth information from sparse RGB-D inputs. Their approach uses a local and global patch-based ray-sampling strategy: global sampling is paired with geometry regularization using warped images and CLIP features, whereas local sampling uses VGG\cite{simonyan2015vgg} features for the perceptual regularization. Explicit depth regularization further guides the geometry, resulting in improved performance compared to previous baselines. Zhong et al.~\cite{zhongEmpoweringSparseInputNeural2025} presented a method that strengthens the NeRF with dual-level semantic guidance from dense novel views. Rendered semantics integrated at both the supervision and feature levels provide robust, augmented data. A bidirectional verification module ensures the reliability of the semantic labels, and a learnable codebook within the MLP encodes semantic-aware patterns. These strategies improve both geometry and appearance modeling by using sparse inputs.

HG3-NeRF~\cite{gaoHG3NeRFHierarchicalGeometric2024} introduces a hierarchical approach that addresses NeRF’s limitations in sparse-view scenarios. The method combines sparse depth priors from SfM for geometric alignment (Hierarchical Geometric Guidance), learns semantic content in a coarse-to-fine manner using CLIP (Hierarchical Semantic Guidance), and applies hierarchical training for appearance consistency. HG3-NeRF achieves high-fidelity synthesis and outperforms state-of-the-art methods on sparse-view benchmark datasets. ViP-NeRF~\cite{somrajViPNeRFVisibilityPrior2023} introduced a visibility prior to regularize NeRF training from sparse views. This method derives a dense visibility prior using plane-sweep volumes and requires no pre-training. It reformulates the NeRF MLP to output visibility, thereby directly reducing computational costs. By combining this visibility prior with sparse depth data, ViP-NeRF achieves state-of-the-art performance for sparse input NeRFs, producing sharper results with fewer artifacts than previous methods.

SimpleNeRF~\cite{somrajSimpleNeRFRegularizingSparse2023} proposes a prior-free regularization approach for few-shot NeRFs. The model is augmented to be biased toward simpler solutions, providing in-situ depth supervision without using external pre-trained depth priors. It applies point augmentation for smoother depth, view augmentation to disable view-dependent radiance, and coarse-fine consistency loss. SimpleNeRF achieves strong performance in both view synthesis and depth estimation. ConsistentNeRF~\cite{huConsistentNeRFEnhancingNeural2023} improves sparse-view NeRF synthesis by enforcing 3D consistency using depth information. It regularizes both multi-view and single-view consistency among pixels. Depth-invariant loss focuses on learning the pixels with reliable 3D correspondences. This led to significant gains in PSNR, SSIM, and LPIPS compared with the standard NeRF baselines.

VGOS~\cite{sunVGOSVoxelGrid2023} adopts a different approach by using voxel grids for radiance-field reconstruction from sparse inputs. The method is trained incrementally to avoid overfitting at the scene periphery and applies a novel color-aware voxel-smoothness loss for regularization. VGOS achieves state-of-the-art results in terms of both quality and speed, converging within 3–5 min without the need for pre-trained models or extra inputs. SANeRF~\cite{xiaoSpatialAnnealingEfficient2024} offers an efficient few-shot neural rendering method by adapting pre-filtering concepts common in hybrid representations. It introduces universal frequency annealing in the spatial domain, making the approach compatible with various NeRF-acceleration methods. SANeRF uses a coarse-to-fine strategy by shrinking the sampling kernel exponentially, achieving high-quality results and superior speed compared to FreeNeRF \cite{yang2023freenerf}.

FrameNeRF~\cite{xingFrameNeRFSimpleEfficient2024} presents a simple three-stage framework for few-shot novel-view synthesis. It uses a regularization model, such as FreeNeRF \cite{yang2023freenerf}, to generate pseudo-dense multiview images from sparse inputs. These synthetic views are used to train a fast, high-fidelity model, such as TensoRF\cite{tensoRF2022}. The model is then fine-tuned using the original sparse views to correct the artifacts and capture realistic details. This approach achieved a state-of-the-art performance. ARC-NeRF~\cite{seoARCNeRFAreaRay2025} introduced Area Ray Casting, a new strategy for few-shot NeRF. Instead of casting single rays, it uses bundles of rays to cover a broader range of unseen views. The method also applies adaptive high-frequency regularization and luminance consistency to improve texture accuracy and training efficiency.

ExtremeNeRF~\cite{leeFewshotNeuralRadiance2023} was the first to address few-shot novel-view synthesis under unconstrained illumination. It uses multiview albedo consistency, geometric alignment, and intrinsic decomposition to address varying lighting conditions. ExtremeNeRF produces sharp and realistic results with fine geometry and sets a new benchmark for in-the-wild datasets. CaesarNeRF~\cite{zhuCaesarNeRFCalibratedSemantic2024} proposes an end-to-end method that combines scene-level-calibrated semantic representations with pixels, explicitly modelling pose differences among reference views and refining the calibration sequentially by aligning viewpoints to precise locations. CaesarNeRF achieves state-of-the-art results for various reference-view counts.

ManifoldNeRF~\cite{kanaokaManifoldNeRFViewdependentImage2023} supervises feature vectors at unknown viewpoints using interpolated features from neighboring known viewpoints. This method enables a more accurate volume representation than those using constant feature vectors. ManifoldNeRF performs well in complex scenes and offers insights into effective viewpoint selection in real-world settings. FrugalNeRF~\cite{linFrugalNeRFFastConvergence2025} offers fast convergence for few-shot novel view synthesis without using learned priors. It shares voxel weights across multiple scales for an efficient and detailed representation. A cross-scale geometric adaptation scheme selects pseudo-ground-truth depths based on reprojection errors to guide the training. FrugalNeRF delivers high quality and significantly reduced training time, making it a practical choice for efficient 3D scene reconstruction.

NeO 360~\cite{irshadNeO360Neural2023} is a generalizable approach for 360\textdegree scene reconstruction from one or a few posed RGB images of a new outdoor environment. It uses a hybrid image-conditional triplanar representation to model complex outdoor 3D scenes. This enables efficient 360 °novel view synthesis, as well as scene editing and composition. NeO 360 significantly outperforms state-of-the-art generalizable NeRF methods on challenging unbounded 360\textdegree datasets. LEAP~\cite{jiangLEAPLiberateSparseview2023} introduced a pose-free strategy for sparse-view 3D modeling. It removes all explicit camera pose operations and learns geometric knowledge directly from image data. LEAP represents each scene as a neural radiance field in a single-feedforward step. The shared neural volume is updated through the feature-similarity-driven aggregation of 2D features. LEAP delivers high-quality novel views from only 2–5 unposed images and matches the quality of pose-based methods while being significantly faster than them.

SparsePose~\cite{sinhaSparsePoseSparseViewCamera2022} addresses the challenge of camera pose estimation in sparse-view scenarios. It recovers accurate poses from fewer than 10 wide-baseline images by learning to regress the initial poses and iteratively refining them using local features and 3D geometric consistency. SparsePose is trained on a large-scale object dataset and outperforms both SfM and other learning-based pose estimation methods, making high-fidelity 3D reconstruction from sparse inputs possible. SparseAGS~\cite{zhaoSparseviewPoseEstimation2024} tackles joint 3D reconstruction and pose estimation for sparse input images, where co-dependent errors can be significant. The framework combines generative priors and outlier reasoning with a combination of discrete searches and continuous optimization. SparseAGS improves both pose accuracy and reconstruction quality compared to baseline methods and demonstrates strong robustness in challenging sparse-view settings.

\begin{table*}[h]
\centering
\scriptsize
\renewcommand{\arraystretch}{1.2}
\rowcolors{2}{rowgray}{white}
\setlength{\tabcolsep}{6pt}
\resizebox{\textwidth}{!}{%
\begin{tabular}{l|c|c|c|c|c}
\toprule
\textbf{Method} & \textbf{Year} & \textbf{Input Views} & \textbf{Pose Needed} & \textbf{Representation Type} & \textbf{Runtime} \\
\midrule
NeRF \cite{mildenhall2020nerf} & 2020 & 100+ & Yes & Radiance Field & Slow \\
BARF \cite{Lin2021BARFBN} & 2021 & 10--20 & Yes & Radiance Field & Slow \\
RegNeRF \cite{niemeyerRegNeRFRegularizingNeural2022} & 2022 & 3--6 & Yes & Radiance Field & Slow (per-scene optimization) \\
pixelNeRF \cite{yu2021pixelnerf} & 2021 & 1--few & Yes & Radiance Field & Feed-forward (Fast inference) \\
IBRNet \cite{wangIBRNetLearningMultiView2021} & 2021 & 8--12 & Yes & Neural Rendering & Efficient (inference) \\
NeRS \cite{zhangNeRSNeuralReflectance} & 2021 & Sparse & Yes & Neural Reflectance Surface & Moderate \\
X-NeRF \cite{zhuXNeRFExplicitNeural2022} & 2022 & Sparse RGB-D & No & Explicit Radiance Field (CNN) & Fast (inference) \\
6Img-to-3D \cite{gieruc6Imgto3DFewImageLargeScale2025} & 2025 & 6 & No & Triplane (Transformer) & Fast \\
MixNeRF \cite{seoMixNeRFModelingRay2023} & 2023 & Sparse & Yes & Radiance Field (Mixture Density) & Efficient \\
PANeRF \cite{ahnPANeRFPseudoviewAugmentation2022} & 2022 & Few-shot & Yes & Radiance Field (Pseudo-views) & Moderate \\
InfoNeRF \cite{kimInfoNeRFRayEntropy2022} & 2022 & Few-shot & Yes & Radiance Field (Prior-free) & Efficient \\
GeCoNeRF \cite{kwakGeCoNeRFFewshotNeural2023} & 2023 & Few-shot (3-5) & Yes & Radiance Field & Fast \\
Putting NeRF on a Diet \cite{jainPuttingNeRFDiet2021} & 2021 & Few-shot (e.g., 8) & Yes & Radiance Field (Semantic Prior) & Efficient \\
DaRF \cite{songDaRFBoostingRadiance2023} & 2023 & Few & Yes & Radiance Field + MDE & Moderate \\
DiViNeT \cite{voraDiViNeT3DReconstruction2023} & 2023 & Sparse & Yes & Neural Surface (Templates) & Moderate \\
HG3-NeRF \cite{gaoHG3NeRFHierarchicalGeometric2024} & 2024 & Sparse & Yes & Radiance Field (Hierarchical Guidance) & Moderate \\
ViP-NeRF \cite{somrajViPNeRFVisibilityPrior2023} & 2023 & Sparse & Yes & Radiance Field (Visibility Prior) & Moderate \\
SimpleNeRF \cite{somrajSimpleNeRFRegularizingSparse2023} & 2023 & Sparse & Yes & Radiance Field (Augmented Models) & Moderate \\
ConsistentNeRF \cite{huConsistentNeRFEnhancingNeural2023} & 2023 & Sparse & Yes & Radiance Field (3D Consistency) & Moderate \\
VGOS \cite{sunVGOSVoxelGrid2023} & 2023 & Sparse (3-10) & Yes & Voxel Grid & Fast (3-5 min training) \\
SANeRF \cite{xiaoSpatialAnnealingEfficient2024} & 2024 & Sparse & Yes & Radiance Field (Spatial Annealing) & Efficient \\
FrameNeRF \cite{xingFrameNeRFSimpleEfficient2024} & 2024 & Sparse & Yes & Radiance Field (Multi-stage) & Moderate \\
ARC-NeRF \cite{seoARCNeRFAreaRay2025} & 2025 & Few-shot & Yes & Radiance Field (Area Ray Casting) & Efficient \\
CaesarNeRF \cite{zhuCaesarNeRFCalibratedSemantic2024} & 2024 & Few-shot & Yes & Radiance Field (Semantic Rep.) & Moderate \\
ManifoldNeRF \cite{kanaokaManifoldNeRFViewdependentImage2023} & 2023 & Few-shot & Yes & Radiance Field (Feature Supervision) & Moderate \\
FrugalNeRF \cite{linFrugalNeRFFastConvergence2025} & 2025 & Extreme few-shot & Yes & Radiance Field (Weight-sharing Voxels) & Fast (10 min training) \\
NeO 360 \cite{irshadNeO360Neural2023} & 2023 & Single/few & Yes & Triplanar (Image-Conditional) & Efficient (generalizable)\\
Preface \cite{buhlerPrefaceDatadrivenVolumetric2023} & 2023 & Few-shot (2) & Yes & Volumetric (NeRF) & Moderate (practical) \\
NeuralLift-360 \cite{xuNeuralLift360LiftingInthewild2023} & 2023 & Single & No & NeRF + Diffusion & Moderate (1.5 hrs training) \\
LEAP \cite{jiangLEAPLiberateSparseview2023} & 2023 & Sparse (2-5) & No & Neural Volume & Fast (0.3s reconstruction) \\
SparsePose \cite{sinhaSparsePoseSparseViewCamera2022} & 2022 & Few wide-baseline ($<$10) & No & Pose Regression & Fast (robust pose estimation) \\
SC-NeuS \cite{huangSCNeuSConsistentNeural2023} & 2023 & Sparse (as few as 3) & No & Neural Surface (SDF) & Moderate \\
\bottomrule
\end{tabular}%
}
\caption{Summary of NeRF-based sparse-view 3D methods: inputs, pose requirements, representations, and reported runtimes. “Sparse” and “few-shot” are defined as 3–10 and 2–5 input views, respectively. Runtime refers to training or inference as specified.}
\label{tab:nerf_comparison}
\end{table*}

SC-NeuS~\cite{huangSCNeuSConsistentNeural2023} introduced a joint learning approach for surface reconstruction with fine details from sparse and noisy camera poses. It directly uses multiview constraints from explicit neural surface geometry, employing fast, differentiable on-surface intersections and view-consistent losses. SC-NeuS consistently outperformed state-of-the-art methods in both surface reconstruction and camera-pose estimation. PoseProbe~\cite{gaoGenericObjectsPose2025} introduced the use of generic objects as “pose probes” for few-shot NeRF reconstruction from 3 to 6 unposed scene images. The method automatically segments a probe object and initializes its shape as a cube. A dual-branch optimization with separate object and scene NeRFs jointly refines the geometry and constrains pose estimation. PoseProbe achieves state-of-the-art performance in both pose estimation and novel view synthesis, particularly in large baseline settings where COLMAP fails.

MixNeRF~\cite{seoMixNeRFModelingRay2023} models each ray with a mixture density for novel view synthesis from sparse inputs. It introduces an auxiliary ray depth estimation task and remodels the colors with new blending weights based on the estimated depth. MixNeRF outperforms other leading methods in terms of both efficiency and quality without relying on external modules or additional supervision. SparseNeuS~\cite{longSparseNeuSFastGeneralizable2022} presents a fast and generalizable method for neural surface reconstruction from sparse-view images. It uses generalizable priors from image features and geometry encoding volumes. SparseNeuS is notable for advancing neural surface reconstruction under sparse conditions.

SPARF~\cite{hamdiSPARFLargeScaleLearning2023} addresses the large-scale learning of 3D sparse radiance fields from few input  data, advancing the understanding of radiance field learning with limited data and providing solutions for both speed and generalization. EG-HumanNeRF~\cite{wang2024eg} introduced an efficient and generalizable human NeRF model that leverages human-specific priors for sparse-view body reconstruction. This approach demonstrates the value of domain priors in improving the efficiency and generalizability. FlexNeRF~\cite{Jayasundara2023FlexNeRFPF} focuses on photorealistic free-viewpoint rendering of moving humans from sparse views. This method adapts the NeRF to handle nonrigid motion, significantly broadening the scope of sparse 3D reconstruction in dynamic real-world scenes. For a comprehensive examination of selected representative methods within the neural implicit paradigm, which emphasizes their fundamental innovations and specific contributions to addressing sparse-view challenges, refer to Table \ref{tab:nerf_comparison}.

In summary, methods like PixelNeRF \cite{yu2021pixelnerf} and IBRNet \cite{wangIBRNetLearningMultiView2021} pioneered generalizable NeRFs, moving beyond per-scene optimization. While ZeroRF\cite{shiZeroRFFastSparse2023} pushed the boundaries of rapid 360° reconstruction, approaches such as 6Img-to-3D \cite{gieruc6Imgto3DFewImageLargeScale2025} demonstrated scalability to large outdoor scenes, highlighting a diversification of focus from pure fidelity to practical deployment considerations. However, they often face computational bottlenecks and challenges in real-time application.

\section{3D Gaussian Splatting Approaches}
\label{sec:3dgs-hybrid}

The introduction of 3DGS~\cite{kerbl2023threedgs} represents a major shift in novel-view synthesis and 3D reconstruction. Unlike implicit methods such as NeRF, 3DGS models a scene explicitly as a collection of 3D Gaussians, allowing for extremely fast training and real-time rendering. However, the original 3DGS approach requires dense input views, and its performance decreases significantly in sparse-view settings, resulting in artifacts and poor geometry. This section reviews the key advancements in 3DGS sparse-view reconstruction. We organized these methods into three main categories: core 3DGS, hybrid approaches that integrate diffusion models, and specialized pose-free methods.

\subsection{Core 3DGS Methods}
\label{sec:pure-3dgs}

This subsection focuses on methods that enhance the core 3DGS framework to improve its performance in sparse-view reconstructions. These advancements primarily stem from innovations in initialization, regularization, and optimization strategies that address the limitations of sparse inputs.

\subsubsection{Initialization and Floater Mitigation}
A critical challenge in sparse-view 3DGS is managing spurious geometry, often termed 'floaters,' and ensuring robust scene initialization with limited data. The methods in this category focus on refining the initial placement and properties of Gaussians.

SparseGS~\cite{xiongSparseGSRealTime360deg2025} is an efficient pipeline that targets common artifacts like "floaters" and "background collapse" in 3DGS trained with few inputs. It introduces new depth rendering methods, such as "mode-selection depth" and "softmax-scaling depth," to guide Gaussian placement and reduce floaters. Depth priors were incorporated using a patch-based depth correlation loss. The Unseen Viewpoint Regularization (UVR) module uses Score Distillation Sampling (SDS)\cite{poole2022dreamfusion} from large vision models, such as Stable Diffusion\cite{rombach2022high}, to guide the training from distant viewpoints, helping to prevent overfitting. SparseGS also features an advanced floater pruning procedure. It achieves state-of-the-art results in 360-degree and forward-facing sparse-view synthesis, improving the quality and reducing artifacts with as few as 3 to 12 input images.

LoopSparseGS~\cite{baoLoopSparseGSLoopBased2024} uses a loop-based 3DGS approach to address problems such as too few initial points, weak supervision, and oversized Gausssians. Its Progressive Gaussian Initialization (PGI) iteratively densifies the point cloud with rendered pseudo-images and real training images. Depth-alignment Regularization (DAR) aligns sparse SfM depth and dense monocular depth using a sliding window approach for better supervision. Sparse-friendly Sampling (SFS) splits large Gaussian ellipsoids based on pixel error, which reduces blurring and overfitting. LoopSparseGS provides high-quality and detailed renderings with efficient training on various datasets. CoR-GS~\cite{zhangCoRGSSparseView3D2024} introduces a co-regularization strategy for sparse-view 3DGS. It simultaneously trains two 3D Gaussian radiance fields using their point and rendering disagreements for self-supervision. Co-pruning removes Gaussians from inaccurate positions, whereas pseudo-view co-regularization suppresses rendering errors by sampling online pseudo-views. The CoR-GS regularizes geometry, produces compact representations, and achieves state-of-the-art novel view synthesis across different datasets using fewer Gaussians.

GaussianObject~\cite{yangGaussianObjectHighQuality3D2024} is designed for high-quality 3D object reconstruction from as few as four views. It adds explicit structure priors and a diffusion-based repair model to 3DGS. The method uses a visual hull to initialize Gaussians and a KNN-based 'floater' removal technique. A "Gaussian repair model" based on a fine-tuned ControlNet\cite{zhang2023adding} corrects problematic Gaussians in poorly observed regions, which are trained using leave-one-out and 3D noise strategies. GaussianObject achieves strong perceptual and quantitative results in object-centric scenes and offers a COLMAP-free option.

\subsubsection{Regularization and Consistency}

Ensuring geometric consistency and preventing overfitting are paramount in sparse-view 3DGS. Researchers have developed various regularization techniques to stabilize training and improve reconstruction fidelity. CoR-GS \cite{zhangCoRGSSparseView3D2024} introduces a unique co-regularization strategy by simultaneously training two 3D Gaussian radiance fields and leveraging their point and rendering disagreements for self-supervision. This approach includes co-pruning to remove inaccurate Gaussian distributions and pseudo-view co-regularization to suppress rendering errors, resulting in more compact and accurate representations.

DNGaussian~\cite{liDNGaussianOptimizingSparseView2024} optimizes sparse-view 3D Gaussian Radiance Fields with Global-Local Depth Normalization and Hard and Soft Depth Regularization. It focuses on restoring scene geometry using coarse monocular depth supervision while preserving fine color details. By freezing the Gaussian shape parameters and centers during specific phases and normalizing the depth globally and locally, DNGaussian mitigates geometry degradation. It achieves state-of-the-art results with training speeds up to 25 times faster than some NeRFs and real-time rendering at 300 frames per second (FPS).

DropGaussian~\cite{parkDropGaussianStructuralRegularization2025} proposes a structural regularization technique for sparse-view 3DGS that randomly removes Gaussians during training. This approach increases the visibility and gradient flow for the remaining Gaussians and helps reduce overfitting. DropGaussian uses a progressively increasing dropping rate and achieves competitive results with prior-based 3DGS, without extra complexity or computational cost.

S2Gaussian~\cite{wanS2GaussianSparseViewSuperResolution2025} targets high-quality 3D reconstruction from sparse, low-resolution input views. The method has two stages: first, it optimizes a low-resolution Gaussian representation and densifies it using Gaussian Shuffle Split. Second, it refines high-resolution Gaussians using super-resolved images and a blur-free inconsistency modeling scheme based on robust 3D optimization. S2Gaussian achieves state-of-the-art results, producing accurate and detailed 3D scenes. SCGaussian~\cite{pengStructureConsistentGaussian2024} enforces 3D-consistent scene structure in few-shot 3DGS using matching priors. It introduces a hybrid Gaussian representation with ordinary and ray-based Gaussians, along with a dual-optimization strategy that constrains the position and shape of each Gaussian. Ray-based Gaussians are bound to matching rays, restricting their optimization and ensuring accurate surface convergence. SCGaussian delivers state-of-the-art rendering quality and efficiency.

UGOT~\cite{sunUncertaintyguidedOptimalTransport2024} (Uncertainty-guided Optimal Transport) optimizes the depth distribution in 3DGS for sparse views by integrating uncertainty estimates from pretrained generative diffusion models. It focuses on training more reliable depth data and reducing overfitting and artifacts from noisy monocular depth. UGOT applies optimal transport to align the sampled depth with the ground truth, achieving superior novel-view synthesis and faster convergence.

\subsubsection{Generalization and Efficiency}
The pursuit of faster training, real-time rendering, and cross-scene generalizability has driven significant advancements in core 3DGS methods. Speedy-Splat~\cite{hansonSpeedySplatFast3D} enhances 3DGS for real-time novel view synthesis by increasing the rendering speed and reducing the model size. It introduces a precise tile Intersect (SnugBox and AccuTile) to localize Gaussians accurately in the image, which reduces unnecessary pixel processing. Efficient Pruning, using both soft and hard pruning, further reduces the number of Gaussians without sacrificing image quality. These techniques boost rendering speed by up to 6.71x and shrink model size by 10.6x, with little loss in quality, making 3DGS suitable for resource-limited settings.

TranSplat~\cite{zhangTranSplatGeneralizable3D} is a generalizable 3DGS method that addresses multi-view feature matching challenges for sparse inputs. It uses a transformer-based architecture with a predicted depth confidence map to guide local feature matching using a Depth-aware Deformable Matching Transformer. Monocular depth estimation models provide prior knowledge using a depth-refined U-Net. This setup enabled the precise estimation of 3D Gaussian centers, even in non-overlapping or low-texture regions. TranSplat achieves state-of-the-art benchmark results, competitive speed, and strong cross-dataset generalization performance. VGNC~\cite{linVGNCReducingOverfitting2025} aims to reduce overfitting in sparse-view 3DGS through validation-guided Gaussian number control. This is the first method to use generative novel view synthesis models, such as ViewCrafter~\cite{yu2024viewcrafter} to create validation images. These images guide a growth-and-dropout mechanism that dynamically adjusts the number of Gaussians, thereby helping identify the optimal count. The VGNC improves the test set rendering quality, reduces memory use, and accelerates both training and rendering, making 3DGS more practical for sparse input.

UniForward~\cite{tianUniForwardUnified3D2025} introduced a feed-forward 3DGS model that unifies 3D scene and semantic field reconstruction from sparse, uncalibrated, and unposed views. It embeds anisotropic semantic features into 3D Gaussians using a dual-branch decoupled decoder. A loss-guided view sampler stabilizes training without ground truth depth or masks. UniForward achieves state-of-the-art performance in real-time novel-view synthesis and view-consistent semantic segmentation, requiring only images as input.

SparSplat~\cite{jenaSparSplatFastMultiView2025} presents a fast and generalizable multiview reconstruction method using 2D Gaussian Splatting. This method achieves state-of-the-art 3D reconstruction and novel view synthesis from sparse, uncalibrated inputs, offering an unprecedented inference speed. Generalizable Human Gaussians~\cite{kwon2024generalizable} introduced a method for sparse-view synthesis of realistic human avatars using generalizable Gaussian splatting. This study extends the application of generalizable techniques to human-specific object categories, producing high-quality avatars from a limited number of input views.

\subsubsection{Geometry-Prioritized and Surface-Aware 3DGS}
Beyond general improvements, a significant line of research focuses on integrating and refining geometric priors within 3DGS to achieve more accurate surface reconstructions and overcome the inherent ambiguities of sparse data. 

Sparse2DGS~\cite{wuSparse2DGSGeometryPrioritizedGaussian} targets geometry-prioritized surface reconstruction from sparse views. It initializes 2D Gaussian Splatting (2DGS) using an MVS point cloud to obtain a dense geometry. The method fixes color and feature optimization to encourage the learning of accurate geometry and uses direct Gaussian primitive regularization (DGPR) with reparameterization-based disk sampling and cross-view feature consistency. A Selective Gaussian Update (SGU) mechanism further refines the MVS-initialized primitives using rendered geometric cues. Sparse2DGS achieves state-of-the-art surface reconstruction accuracy and is significantly faster than NeRF-based fine tuning.

HiSplat~\cite{tangHiSplatHierarchical3D2024} presents a hierarchical 3DGS for sparse-view reconstruction, particularly in difficult two-view cases. It uses a coarse-to-fine approach to build 3D Gaussian models that capture both the broad structures and fine textures. Key modules include an Error Aware Module (EAM) for Gaussian compensation and a Modulating Fusion Module (MFM) for Gaussian repair. These foster important inter-scale interactions, leading to top performance in novel view synthesis and strong cross-dataset generalization capability.

PointGS~\cite{xiangPointGSPointAttentionAware2025} advances 3DGS for sparse-view synthesis with a multi-pronged strategy. It begins with dense initialization from a stereo foundation model (VGGT)\cite{wang2025vggt} for accurate camera poses and dense point clouds. The method aggregates multi-scale 2D appearance features and uses a self-attention network for point-wise interactions. With added depth regularization, PointGS surpasses NeRF-based methods and matches leading 3DGS approaches in few-shot settings while preserving the details and minimizing artifacts.

Chan et al.~\cite{chanPointCloudDensification2024} improve sparse-view 3DGS by prioritizing robust point cloud initialization over standard depth-based regularization, which can be error-prone. They introduced Systematically Angle of View Sampling (SAOVS) for better side-view coverage and applied semantic pseudo-label regularization to guide the reconstruction. This method consistently outperformed the standard 3DGS baselines on datasets such as ScanNet and LLFF, yielding high-quality, novel views with minimal distortion.  Kim et al.~\cite{kimImprovingGeometrySparseView2024} address geometric degradation in sparse-view 3DGS by reparameterizing Gaussian positions according to uncertainty. This method separates low-uncertainty image-plane-parallel DoFs from high-uncertainty ray-aligned DoFs and applies the targeted constraints. Bounded offset and visibility loss terms are used to reduce artifacts, resulting in visually coherent and geometrically accurate reconstructions, even with very limited data.

SPARS3R~\cite{tangSPARS3RSemanticPrior} combines accurate pose estimation from SfM with dense point clouds from modern depth techniques, such as DUSt3R\cite{Wang2023DUSt3RG3} and MASt3R\cite{li2023mast3r}. It uses a two-stage alignment: Global Fusion Alignment for coarse alignment, followed by Semantic Outlier Alignment to refine regions with depth discrepancies using semantic segmentation. This creates a dense, pose-accurate 3D prior for Gaussian optimization, leading to photorealistic rendering from sparse images and previous SfM-based initialization methods on the DTU and LLFF datasets. 

CoMapGS~\cite{jangCoMapGSCovisibilityMapbased2025} reframes sparse view synthesis using pixel-wise covisibility maps for adaptive supervision and initial-point-cloud enhancement. This approach improves the initialization in both the multiview and monoview regions using a covisibility map-based weighting to target region-wise imbalances. CoMapGS effectively recovers high-uncertainty regions, leading to strong overall performance.

FewViewGS~\cite{yinFewViewGSGaussianSplatting2024} improves 3DGS under sparse-view conditions without external priors. It uses a multistage training scheme with matching-based consistency constraints applied to the novel views. These constraints match features from training images to supervise novel views using color, geometric, and semantic losses. Locality-preserving regularization helps to remove artifacts, yielding more reliable renderings. SolidGS~\cite{shenSolidGSConsolidatingGaussian2024} addresses sparse-view surface reconstruction by consolidating Gaussians with a generalized exponential Gaussian distribution and by adding new geometric constraints. A global learnable solidness factor makes Gaussians more opaque and reduces multi-view depth inconsistencies. An additional self-supervised geometry loss from virtual views and monocular normal estimation guided the optimization. SolidGS achieves state-of-the-art geometry and novel view synthesis quality.

MVPGS~\cite{xuMVPGSExcavatingMultiview2024} introduced a new approach for few-shot novel view synthesis using 3DGS, leveraging geometric priors from Multi-View Stereo (MVS). This method uses a learning-based MVS for strong geometric initialization and applies a forward warping technique to impose appearance constraints. The MVPGS also adds view-consistent geometry constraints for the Gaussian parameters and uses monocular depth regularization. It achieves state-of-the-art performance with real-time rendering. GeoRGS~\cite{liuGeoRGSGeometricRegularization2024} is a prior-independent 3DGS method that corrects erroneous Gaussian growth and addresses depth distortion for sparse-input cases. It introduces Seed-based Geometric Regularization (S.G.R) to guide the growth of Gaussians and ensure accurate scene geometry. Depth smoothness and consistency regularization terms further align the reconstructions with the real-world geometry, resulting in top performance and high efficiency.

Chung et al.~\cite{chungDepthRegularizedOptimization3D2024} proposed a depth-regularized optimization for 3DGS from few-shot images. They used a dense depth map as a geometric guide to avoid overfitting. The scale and offset of dense depth maps were refined using sparse COLMAP feature points, which enforced geometric constraints during color-based optimization. This approach enables a plausible geometry and visually attractive results with very few input images. MVG-splatting ~\cite{Li2024MVGSplattingMG} introduced Multi-View Guided Gaussian Splatting with adaptive quantile-based geometric consistency, which improves multi-view consistency and densification in 3DGS, resulting in more robust and accurate reconstructions from sparse inputs.

InfoNorm~\cite{wang2024infonorm} presents a mutual information-based approach for shaping surface normals in sparse-view reconstruction. By leveraging information theory, InfoNorm enhances the geometric accuracy and consistency under challenging sparse input conditions. UniGS~\cite{Wu2024UniGS} proposed a model for novel view synthesis and 3D reconstruction that predicts high-fidelity 3D Gaussians from any number of posed, sparse-view images. Unlike methods that regress Gaussians per pixel and concatenate them, often causing "ghosting," UniGS models unitary 3D Gaussians directly in the world space. It updates these Gaussian layers by layer with a DETR-like framework \cite{carion2020end} using multi-view cross-attention (MVDFA), effectively avoiding ghosting and allocating more Gaussians to complex regions. UniGS supports variable input view counts without requiring retraining.A detailed summary of representative 3D Gaussian Splatting (3DGS)-based methods for sparse-view 3D reconstruction can be found in Table~\ref{tab:3dgs_comparison}.

\begin{table*}[h]
\centering
\scriptsize
\renewcommand{\arraystretch}{1.2}
\rowcolors{2}{rowgray}{white}
\setlength{\tabcolsep}{6pt}
\resizebox{\textwidth}{!}{%
\begin{tabular}{l|c|c|c|c|c}
\toprule
\textbf{Method} & \textbf{Year} & \textbf{Input Views} & \textbf{Pose Needed} & \textbf{Representation Type} & \textbf{Runtime} \\
\midrule
3DGS \cite{kerbl2023threedgs} & 2023 & Dense & Yes & 3DGS & Real-time (rendering) \\
CF-3DGS \cite{fuCOLMAPFree3DGaussian} & 2024 & Video & No & 3DGS & Fast (training) \\
InstantSplat \cite{fanInstantSplatUnboundedSparseview} & 2024 & 6--12 & No & 3DGS + Dense Stereo & Fast (reconstruction $<$1 min)\\
SparseGS \cite{xiongSparseGSRealTime360deg2025} & 2025 & 3--12 & Yes & 3DGS & Moderate \\
Intern-GS \cite{sunInternGSVisionModel2025} & 2025 & Sparse & No & 3DGS + VFM & Fast \\
Speedy-Splat \cite{hansonSpeedySplatFast3D} & 2024 & Sparse & Yes & 3DGS (Pruned) & Fast (rendering) \\
DropGaussian \cite{parkDropGaussianStructuralRegularization2025} & 2025 & Sparse (as few as 3) & Yes & 3DGS (Structural Reg.) & Competitive \\
S2Gaussian \cite{wanS2GaussianSparseViewSuperResolution2025} & 2025 & Sparse, low-res & Yes & 3DGS (Super-Resolution) & Moderate \\
SCGaussian \cite{pengStructureConsistentGaussian2024} & 2024 & Few-shot & Yes & 3DGS (Hybrid/Ray-based) & Efficient \\
CoMapGS \cite{jangCoMapGSCovisibilityMapbased2025} & 2025 & Sparse & Yes & 3DGS (Covisibility Map) & Efficient \\
FewViewGS \cite{yinFewViewGSGaussianSplatting2024} & 2024 & Sparse & Yes & 3DGS (Multi-stage) & Efficient \\
SolidGS \cite{shenSolidGSConsolidatingGaussian2024} & 2024 & Sparse & Yes & 3DGS (Surfel Splatting) & Fast (3 min training) \\
MVPGS \cite{xuMVPGSExcavatingMultiview2024} & 2024 & Few-shot & Yes & 3DGS (MVS Priors) & Real-time \\
GeoRGS \cite{liuGeoRGSGeometricRegularization2024} & 2024 & Sparse & Yes & 3DGS (Geometric Reg.) & Fast \\
pixelSplat \cite{charatanPixelSplat3DGaussian2024} & 2024 & Image pairs & Yes & 3DGS (Probabilistic Depth) & Real-time (rendering) \\

UniGS \cite{Wu2024UniGS} & 2025 & Arbitrary sparse & Yes & Unitary 3D Gaussians (DETR-like) & Fast \\
SparSplat \cite{jenaSparSplatFastMultiView2025} & 2025 & Sparse & No & 2DGS (Generalizable) & Very Fast (inference) \\
\bottomrule
\end{tabular}%
}
\caption{Comparison of 3DGS-based methods for sparse-view 3D reconstruction. “Sparse” and “few-shot” refer to inputs with 2–12 images, respectively. Runtime describes the reported training or inference speed.}
\label{tab:3dgs_comparison}
\end{table*}

\subsection{3DGS + Diffusion Hybrids}
\label{sec:3dgs-diffusion-hybrids}

This category includes methods that combine the generative power of diffusion models with the efficiency of 3DGS to improve sparse-view reconstruction, particularly for hallucinating missing details and maintaining a visual consistency.

Deceptive-NeRF/3DGS~\cite{liuDeceptiveNeRF3DGSDiffusionGenerated2024} enhances sparse-view reconstruction by generating high-quality, photorealistic pseudo-observations with a specialized deceptive diffusion model. Instead of acting as a simple regularizer, this diffusion model serves as a "View Densifier," expanding the sparse dataset by 5 to 10 times. This leads to better reconstruction quality, faster training, and enables super-resolution novel view synthesis, even from poor initial data. Wang et al.~\cite{wangHowUseDiffusion2024} addressed the shortcomings of traditional Score Distillation Sampling (SDS)\cite{poole2022dreamfusion} with Inline Prior Guided Score Matching (IPSM). The IPSM uses visual inline priors from warped views to correct the distribution inconsistencies in the rendered images. Built on a 3DGS backbone, IPSM-Gaussian also adds depth and geometry consistency regularization. This approach achieves state-of-the-art visual fidelity and geometric accuracy, particularly in sparse scenarios.

LM-Gaussian~\cite{yuLMGaussianBoostSparseview2024} improves sparse-view 3DGS using priors from large-scale vision models. It features a background-aware depth-guided initialization for robust point-cloud and accurate poses. Multimodal regularized Gaussian reconstruction uses depth, normal, and virtual-view constraints to avoid overfitting. The iterative Gaussian refinement and scene enhancement modules utilize image and video diffusion priors to further improve the scene details and visual consistency. LM-Gaussian achieves strong 360-degree reconstruction quality with limited input.

MVSplat360~\cite{chenMVSplat360FeedForward3602024} offers a feedforward solution for 360° novel view synthesis using as few as five input images. It combines a geometry-aware 3DGS for coarse reconstruction with a pretrained Stable Video Diffusion (SVD) model\cite{blattmann2023stablevideo} for appearance refinement. By rendering features directly in the SVD latent space, end-to-end training is enabled, boosting both the visual quality and 3D consistency. MVSplat360 sets new benchmarks for wide-sweeping and 360\textdegree NVS tasks, particularly in complex real-world scenes. ProSplat~\cite{luProSplatImprovedFeedForward2025} is a two-stage feedforward framework that improves 3DGS performance on wide-baseline sparse views. It first uses a 3DGS generator and then refines the results using a one-step diffusion-based improvement model. Key features include Maximum Overlap Reference view injection (MORI) for enhancing texture and color and Distance-Weighted Epipolar Attention (DWEA) for geometric consistency. ProSplat consistently outperformed state-of-the-art methods in terms of PSNR on challenging datasets, delivering robust and efficient results for immersive media.

FlowR~\cite{fischerFlowRFlowingSparse2025} narrows the quality gap between sparse and dense 3D reconstructions by using a multiview flow-matching model. It learns a velocity field to align incorrect novel view renderings from sparse reconstructions with ground-truth images, refining them for higher fidelity. FlowR typically relies on 3DGS for initial reconstructions and then applies flow-based refinements. This approach yields consistent, sharp outputs and surpasses previous methods on multiple NVS benchmarks. AugGS~\cite{duAugGSSelfaugmentedGaussians2024} is a two-stage Gaussian-splatting method for sparse-view 3D reconstruction. It uses self-augmented data from a fine-tuned 2D diffusion model and incorporates structural masks. The first stage creates a basic 3DGS representation, and the second stage refines the Gaussian attributes using pseudo-labels generated by a fine-tuned ControlNet\cite{zhang2023adding}. Structural masks further improved robustness. AugGS achieves state-of-the-art perceptual quality and multiview consistency from few inputs, and delivers notable training and inference efficiency.

V3D~\cite{chenV3DVideoDiffusion2024} reconceptualizes 3D generation by treating dense multi-view synthesis as a video generation problem and leveraging pretrained video diffusion models for spatiotemporal consistency. These models are fine-tuned on 3D datasets and integrated with 3D reconstruction pipelines, such as 3DGS, to produce high-quality 3D objects or scenes in minutes. V3D enforces geometric consistency priors and achieves superior object-centric and scene-level novel view synthesis. CAT3D~\cite{gaoCAT3DCreateAnything2024} efficiently generates 3D scenes and objects from limited inputs, such as a single image or text, by decoupling the generation from reconstruction. It uses a multiview diffusion model based on \cite{ho2022video} with 3D self-attention to synthesize consistent novel views, which are then processed by a robust 3D reconstruction pipeline, often 3DGS-based. CAT3D enables rapid and high-quality 3D content creation, making 3D generation more accessible.

CAT4D~\cite{wuCAT4DCreateAnything2024} extends multiview diffusion to 4D, thereby enabling dynamic 3D scene reconstruction from monocular videos. It uses a multiview video diffusion model to transform a single video into a consistent multiview sequence. These sequences are then used to reconstruct a deformable 3D Gaussian representation of the dynamic scene, all without requiring synchronized multi-view capture or additional supervision. RI3D~\cite{paliwalRI3DFewShotGaussian2025} introduced a 3DGS-based approach that separates view synthesis into the reconstruction of visible regions and hallucination of the missing regions. It employs two personalized diffusion models, one for repairing visible areas and another for inpainting missing parts within a two-stage optimization. RI3D produces high-quality textures in occluded or missing regions and outperforms the state-of-the-art methods.

latentSplat~\cite{wewerLatentSplatAutoencodingVariational2024} combines regression-based modeling with a lightweight generative approach for generalizable 3D reconstruction. It uses variational 3D Gaussians to explicitly model uncertainty by assigning distributions of semantic features to predicted 3D locations. latentSplat achieves state-of-the-art results in two-view reconstruction and generalization, particularly with wide input baselines and view extrapolation, while maintaining fast and scalable inference. Chen et al.~\cite{Chen2025ImprovingNVS} present a framework to improve novel view synthesis of 360° scenes from extremely sparse views. DUSt3R\cite{Wang2023DUSt3RG3} was used for camera pose estimation and dense point-cloud generation. Additional views were densely sampled from the upper hemisphere, rendered as synthetic images, and enhanced using a retrained diffusion-based model. Training the 3DGS on these reference and synthetic images expands the scene coverage and reduces overfitting, significantly improving the quality of the extremely sparse inputs.

3D Gaussian Splatting methods excel in rendering speed and offer competitive reconstruction quality for sparse input data. Their explicit nature allows for the effective integration of priors and pose-free optimization, marking a shift towards practical real-time systems.

\subsection{Pose-Free Methods}
\label{sec:pose-free-3dgs}

A key challenge in sparse-view 3D reconstruction is the need for accurate camera poses, which are typically estimated using slow or unreliable SfM pipelines. Pose-free methods address this issue by removing the dependency on external pose estimation, making the reconstruction process more robust and practical for real-world uncalibrated scenarios. Recent pose-free techniques often rely on 3DGS because of its explicit representation of scenes.

InstantSplat~\cite{fanInstantSplatUnboundedSparseview} provides unbounded pose-free Gaussian splatting in just 40 seconds. It uses dense stereo models (DUSt3R)~\cite{Wang2023DUSt3RG3} for coarse geometric initialization and a fast 3D-Gaussian optimization (F-3DGO) module that jointly optimizes the 3D Gaussian attributes and camera poses. This system rapidly reconstructs large-scale scenes and produces high-quality view synthesis from sparse, unposed images in less than one minute, delivering strong rendering quality and pose accuracy. CF-3DGS~\cite{fuCOLMAPFree3DGaussian} enables high-quality novel view synthesis and robust pose estimation without relying on pre-computed parameters from SfM libraries such as COLMAP. By using 3DGS’s explicit point cloud representation and leveraging the temporal continuity in video streams, CF-3DGS sequentially processes the input frames and grows a global 3D Gaussian set. It surpasses prior methods in both view synthesis and pose estimation, particularly with large motions, and provides faster training.

FreeSplatter~\cite{xuFreeSplatterPosefreeGaussian2024} is a pose-free 3DGS framework that creates high-quality 3D Gaussian models and recovers camera parameters from uncalibrated, sparse-view images within seconds. It uses a transformer architecture for multiview information exchange and decodes it into pixel-wise 3D Gaussian primitives. FreeSplatter is scalable and offers strong reconstruction quality and pose estimation, making it well-suited for content creation. MV-DUSt3R+~\cite{tangMVDUSt3RSingleStageScene} is a single-stage feedforward network for dense 3D reconstruction from sparse, unposed RGB images. Unlike pairwise methods, it processes multiple views together and integrates Gaussian splatting heads to regress the 3D Gaussian attributes for synthesizing novel views. MV-DUSt3R+ achieves fast inference, producing dense point clouds and camera poses in less than 2 seconds while improving the reconstruction quality across diverse scenes and view counts.

Gaussian Scenes~\cite{Paul2024GaussianSP} provides a generative and pose-free approach for reconstructing 360-degree scenes from sparse 2D images. It uses depth-enhanced diffusion priors and a new confidence measure for 3D Gaussian Splatting. A diffusion-based generative model inpaints missing details and removes artifacts from novel-view renders and depth maps. These refined views were progressively integrated to achieve multiview consistency. GScenes reconstructs complex 360-degree scenes from pose-free inputs in approximately five minutes. iFusion~\cite{wuIFusionInvertingDiffusion2023} introduced a 3D object reconstruction framework that requires only two views with unknown camera poses. It uses a pretrained novel-view synthesis diffusion model for pose estimation. The model is then fine-tuned for novel view synthesis of the target object, and the registered views with the fine-tuned model are used for 3D reconstruction. Although not strictly a 3DGS method, iFusion’s pose-free design and use of diffusion models make it a relevant approach.

Zhang et al.~\cite{zhangSeeing3DWorld2025} present a snapshot imaging technique for 3D reconstruction of miniature scenes using multi-view images captured with a catadioptric system. This method employs a modified 3D Gaussian Splatting representation enhanced with a visual hull-based depth constraint to handle sparse inputs. Using pre-calibrated virtual cameras, it operates in a pose-free manner without external SfM and achieves state-of-the-art results on miniature scene benchmarks. 

In summary, 3D Gaussian Splatting has rapidly become a leading technique for sparse-view 3D reconstruction. Through advances in core 3DGS methods, hybrids with diffusion models, and robust pose-free strategies, researchers have significantly improved their ability to create high-quality, real-time, and geometrically consistent 3D representations from limited uncalibrated images. These developments have raised the standards for fidelity and practicality in real-world 3D applications.

\section{Integration of Diffusion Models and Vision Foundation Models}
\label{sec:diffusion-vfm}

In sparse-view settings, traditional methods often fail to generate plausible content in the unobserved regions. Recent advances in generative AI, especially diffusion models\cite{ho2020ddpm,ho2022video}, and the rise of Vision Foundation Models (VFMs) such as CLIP\cite{radford2021clip}, SAM\cite{kirillov2023sam}, and DINO\cite{caron2021dino}, are transforming this area~\cite{wangDiffusionModels3D2025}. These models use knowledge acquired from large datasets to synthesize missing details, provide strong priors, and enhance multi-view consistency. This subsection reviews the role of generative diffusion models and Vision Foundation Models, which have become pivotal for overcoming data sparsity by hallucinating missing details and providing semantic priors.

\subsection{Diffusion Models for Enhanced Generation and Consistency}
\label{ssec:diffusion-generation}

Diffusion models were originally developed for 2D generative tasks~\cite{ho2020ddpm} but have since demonstrated strong performance in 3D vision. They are effective in generating high-quality images, synthesizing unseen views, and refining degraded reconstructions, which helps to address the challenges of sparse input data.

GenFusion~\cite{wuGenFusionClosingLoop2025} targets the "conditioning gap" between 3D reconstruction and 3D generation. It introduces a reconstruction-driven video diffusion model that learns to condition video frames on artifact-prone RGB-D renderings. This method uses a cyclical fusion pipeline that progressively adds the restoration frames from the generative model to the training set. This enables progressive expansion and addresses viewpoint saturation. The results show that GenFusion achieves performance on sparse-view datasets comparable to state-of-the-art NeRFs, demonstrating the effectiveness of Gaussian Splatting in these settings. SIR-DIFF~\cite{maoSIRDIFFSparseImage} enhances sparse image sets using a multiview diffusion model. It improves the quality of 2D image collections before 3D reconstruction by filling in missing details and increasing the consistency. This preprocessing step improves the performance of downstream tasks such as 3D reconstruction, feature matching, and depth estimation.

Sp2360~\cite{paulSp2360Sparseview3602024} addresses 360-degree scene reconstruction from sparse views using cascaded 2D diffusion.  It synthesizes new views by inpainting missing regions and eliminating artifacts, and then iteratively adds these views to the training set. This approach achieves multiview consistency and can reconstruct full 360-degree scenes from as few as nine input images. VI3DRM~\cite{chenVI3DRMMeticulous3D2024} presents a diffusion-based model for sparse-view 3D reconstruction. It operates in an ID-consistent and perspective-disentangled 3D latent space, separating semantic information, color, material, and lighting. The model combines real and synthesized images to construct accurate point maps, producing finely textured mesh or point clouds. The VI3DRM delivers highly realistic images and outperforms previous methods on novel-view synthesis benchmarks.

Sparse3D~\cite{zouSparse3DDistillingMultiviewConsistent2024} introduced a 3D reconstruction method for extremely sparse views by distilling robust priors from multiview-consistent diffusion models. This approach uses a controller to extract epipolar features from the input views, guiding a pretrained diffusion model to generate novel-view images that remain 3D-consistent. By leveraging strong 2D priors, Sparse3D produces high-quality novel view synthesis and geometric reconstruction. It also addresses the blurriness common with Score Distillation Sampling (SDS) by introducing category-score distillation sampling. Mao et al.~\cite{maoGeneratingMaterialAware3D} propose a method for creating material-aware, relightable 3D models from sparse views by combining generative diffusion models with an efficient rendering framework. This method factorizes the scene into a differentiable environment illumination model, spatially varying material field, and implicit signed distance function (SDF) field. This enables separate control over geometry, material, and lighting. Mixed supervision using both real and diffusion-generated views improves view consistency, whereas the view selection mechanism filters poor-quality samples for better reconstruction.

ReconFusion~\cite{wuReconFusion3DReconstruction2023} leverages a diffusion prior for novel view synthesis to reconstruct real-world scenes from only a few photos. Trained on synthetic and multi-view datasets, the diffusion prior regularizes a NeRF-based pipeline at novel camera poses beyond the available input. ReconFusion produces realistic geometry and texture in under-constrained regions, significantly outperforming previous few-view NeRF methods.

ReconX~\cite{liuReconXReconstructAny2025} introduced a new 3D scene reconstruction paradigm that frames ambiguous reconstruction as a temporal-generation task. It leverages large pretrained video diffusion models to generate additional observations for the sparse-view reconstruction. ReconX builds a global point cloud, encodes it as a 3D structural condition, and guides the video diffusion model to synthesize 3D-consistent frames. A confidence-aware 3DGS optimization then recovers the scene, achieving state-of-the-art quality and strong generalizability. Zhong et al.~\cite{zhongTamingVideoDiffusion2025} propose a reconstruction-by-generation pipeline for sparse-input 3DGS that utilizes video diffusion models. Their key innovation is a training-free scene-grounding guidance mechanism derived from rendered sequences of an optimized 3DGS model. This mechanism ensures that the video diffusion model generates consistent and plausible sequences, effectively addressing extrapolation and occlusion challenges in sparse-input reconstruction.

MVDiffusion++~\cite{tang2024mvdiffusion++} advances multiview diffusion modeling for single or sparse-view 3D object reconstruction. Its pose-free architecture and view dropout strategy enable dense and high-resolution view synthesis and robust 3D reconstruction from minimal input, providing superior flexibility and scalability. ID-Pose~\cite{cheng2023id} presented a method for sparse-view camera pose estimation by inverting the diffusion models. This generative approach infers camera parameters from limited views, thereby offering a novel solution for pose-free reconstruction pipelines.

Tang et al.~\cite{Tang2024FinetuningTD} present a method to improve sparse-view 3D reconstruction by fine-tuning a pre-trained diffusion model to produce "3D-aware images." This approach uses coarse renderings as image conditions and text prompts as text conditions for the model. A key innovation is the "semantic switch," a self-evaluation mechanism that filters out generated images that do not match the real scenes. This ensures that only informative priors are distilled into the downstream 3D model (such as Instant-NGP)\cite{mueller2022instant}, achieving competitive results with improved cost efficiency.

\subsection{Integration with Vision Foundation Models (VFMs)}
\label{ssec:vfm-integration}

Beyond diffusion, large-scale pretrained Vision Foundation Models (VFMs) provide rich semantic and visual understanding that can significantly enhance sparse-view 3D reconstruction. Trained on vast datasets for various 2D vision tasks, VFMs offer strong priors for segmentation, feature extraction, and image completion.

Integrating VFMs such as CLIP~\cite{radford2021clip}, SAM~\cite{kirillov2023sam}, and DINO~\cite{caron2021dino} is foundational to many state-of-the-art methods. For example, CLIP embeddings can guide semantic supervision, whereas SAM provides object masks or visual hulls~\cite{yangGaussianObjectHighQuality3D2024}. DINO features support robust matching and regularization. These integrations act as powerful priors, supporting initialization, guiding optimization, and generating augmented data when explicit three-dimensional (3D) information is limited. By injecting high-level semantic and perceptual information, VFMs help resolve ambiguities in sparse multiview data and improve both the geometric accuracy and visual fidelity in novel views.

Overall, the use of diffusion models and VFMs marks a new frontier in sparse-view 3D reconstruction research. These generative and semantic models infer unseen data, enforce consistency, and enhance reconstructions with detailed semantics and appearances, directly addressing the challenges of limited observations. A detailed comparison of representative diffusion and hybrid methods for sparse-view 3D is provided in Table~\ref{tab:diffusion_hybrid_comparison}.

\begin{table*}[h]
\centering
\scriptsize
\renewcommand{\arraystretch}{1.2}
\rowcolors{2}{rowgray}{white}
\setlength{\tabcolsep}{6pt}
\resizebox{\textwidth}{!}{%
\begin{tabular}{l|c|c|c|c|c}
\toprule
\textbf{Method} & \textbf{Year} & \textbf{Input Views} & \textbf{Pose Needed} & \textbf{Representation Type} & \textbf{Runtime} \\
\midrule
Deceptive-NeRF/3DGS \cite{liuDeceptiveNeRF3DGSDiffusionGenerated2024} & 2024 & Sparse & Yes & NeRF/3DGS + Diffusion & Efficient (faster training) \\
Sp2360 \cite{paulSp2360Sparseview3602024} & 2024 & 9+ & No & 3DGS + Diffusion & Fast \\
MatSparse3D \cite{maoGeneratingMaterialAware3D} & 2024 & 5 & Yes & Neural Surface + Diffusion & Fast \\
Sparse3D \cite{zouSparse3DDistillingMultiviewConsistent2024} & 2024 & 2--3 & Yes & NeRF + Diffusion & Moderate \\
GenFusion \cite{wuGenFusionClosingLoop2025} & 2025 & Sparse & Yes & GS + Video Diffusion & Moderate (denoising steps increase time) \\
VI3DRM \cite{chenVI3DRMMeticulous3D2024} & 2024 & 4 & Yes & Diffusion-based (Latent Space) & Fast \\
Fine-tuning Diffusion Model \cite{tangFinetuningDiffusionModel2024} & 2024 & Few-shot & Yes & Diffusion-based (NeRF opt.) & Efficient \\
SIR-DIFF \cite{maoSIRDIFFSparseImage} & 2024 & sparse & No & Multi-view Diffusion & Fast \\
iFusion \cite{wuIFusionInvertingDiffusion2023} & 2023 & 2 & No & Diffusion-based & Moderate \\
How to Use Diffusion Priors for Sparse View Synthesis \cite{Wang2024HowTU} & 2024 & Sparse & Yes & 3DGS + Diffusion & Moderate \\
SparseFusion \cite{zhouSparseFusionDistillingViewConditioned2023} & 2023 & 2 & Yes & NeRF + Diffusion & Moderate \\
V3D \cite{chenV3DVideoDiffusion2024} & 2024 & Single image or monocular video & Yes & Video Diffusion + 3DGS/Mesh & Fast \\
CAT3D \cite{gaoCAT3DCreateAnything2024} & 2024 & Single/few images or text & Yes & Multi-view Diffusion & Fast ($<$1 min) \\
AugGS \cite{duAugGSSelfaugmentedGaussians2024} & 2024 & 4--9 & Yes & 3DGS + Diffusion & Fast \\
ProSplat \cite{luProSplatImprovedFeedForward2025} & 2025 & Sparse, wide-baseline & Yes & 3DGS + Diffusion & Moderate \\
Gaussian Scenes \cite{Paul2024GaussianSP} & 2025 & Sparse & No & 3DGS + Diffusion & Fast \\
GaussianObject \cite{yangGaussianObjectHighQuality3D2024} & 2024 & 4 & Yes/No & 3DGS + Diffusion & Fast \\
FlowR \cite{fischerFlowRFlowingSparse2025} & 2025 & Sparse/Dense & Yes & 3DGS + Flow Matching & Moderate \\
RI3D \cite{paliwalRI3DFewShotGaussian2025} & 2025 & Extremely sparse (3-9) & Yes & 3DGS + Diffusion & Moderate \\
latentSplat \cite{wewerLatentSplatAutoencodingVariational2024} & 2024 & 2 video frames & Yes & Variational 3D Gaussians + Generative Decoder & Fast \\
\bottomrule
\end{tabular}%
}
\caption{Summary of diffusion and hybrid approaches for sparse-view 3D reconstruction. “Sparse” refers to input sets with fewer than 10 images; runtimes are qualitative, with “Fast” indicating less than 1 minute per frame and “Moderate” indicating 1–10 minutes per frame.}
\label{tab:diffusion_hybrid_comparison}
\end{table*}

\begin{table*}[p]
\centering
\scriptsize
\rowcolors{2}{rowgray}{white}
\resizebox{\textwidth}{!}{%
\begin{tabular}{lccccccccccc}
\toprule
\textbf{Method} & \textbf{DTU} & \textbf{LLFF/} & \textbf{Tanks \&} & \textbf{CO3D} & \textbf{RealEstate10K/} & \textbf{ScanNet/} & \textbf{MVImgNet} & \textbf{ShapeNet} & \textbf{OpenIllum} & \textbf{ACID} \\
& & \textbf{Mip-NeRF 360} & \textbf{Temples} & & \textbf{DL3DV} & \textbf{ScanNet++} & & & & \\
\midrule
\multicolumn{11}{l}{\textbf{Geometry-Based Methods}} \\
\midrule

Stereo Radiance Fields (SRF) \cite{chibaneStereoRadianceFields2021} & $\checkmark$ & $\checkmark$ & & & & & & & & \\
NOPE-SAC \cite{tanNOPESACNeuralOnePlane2023} & & & & & & $\checkmark$ & & & & \\
Neural 3D reconstruction... \cite{muNeural3DReconstruction2023} & $\checkmark$ & & & & & & & & & \\
A Semantically Aware... \cite{weiSemanticallyAwareMultiView2024} & & & & & & & & & & \\
SparseCraft \cite{younesSparseCraftFewShotNeural2024} & $\checkmark$ & & & & & & & & & \\
3DFIRES \cite{jin3DFIRESFewImage2024} & & & & & & & & & & \\
The Less You Depend... \cite{wangLessYouDepend2025} & & & & & & & & & & \\
Dust to Tower \cite{caiDustTowerCoarsetoFine2024} & & & & & & & & & & \\
SparseAGS \cite{zhaoSparseviewPoseEstimation2024} & & & & & & & & & & \\
SpaRP \cite{xuSpaRPFast3D2024} & & & & & & & & & & \\
sshELF \cite{najafliSshELFSingleShotHierarchical2025} & & & & & & & & & & \\
3D Vessel Reconstruction... \cite{liu3DVesselReconstruction2024} & & & & & & & & & & \\
GS4 \cite{jiangGS4GeneralizableSparse2025} & & & & & & $\checkmark$ & & & & \\
\midrule
\multicolumn{11}{l}{\textbf{Neural Implicit Representations (NeRF and Variants)}} \\
\midrule
NeRF \cite{mildenhall2020nerf} & & $\checkmark$ & & & & & & & & \\
pixelNeRF \cite{yu2021pixelnerf} & $\checkmark$ & & & & & & & $\checkmark$ & & \\
IBRNet \cite{wangIBRNetLearningMultiView2021} & & & & & & & & & & \\
NeRS \cite{zhangNeRSNeuralReflectance} & & & & & & & & & & \\
RegNeRF \cite{niemeyerRegNeRFRegularizingNeural2022} & $\checkmark$ & $\checkmark$ & & & & & & & & \\
Dense Depth Priors... \cite{roessleDenseDepthPriors2022} & & & & & & & & & & \\
DS-NeRF \cite{dengDepthsupervisedNeRFFewer2022} & & & & & & & & & & \\
X-NeRF \cite{zhuXNeRFExplicitNeural2022} & & $\checkmark$ & & & & & & & & \\
SparseNeuS \cite{longSparseNeuSFastGeneralizable2022} & $\checkmark$ & & & & & & & & & \\
SparseNeRF \cite{wangSparseNeRFDistillingDepth2023} & & & & & & & & & & \\
FlipNeRF \cite{seoFlipNeRFFlippedReflection2023} & $\checkmark$ & & & & & & & & & \\
ZeroRF \cite{shiZeroRFFastSparse2023} & & $\checkmark$ & & & & & & & $\checkmark$ & \\
SPARF \cite{hamdiSPARFLargeScaleLearning2023} & & & & & & & & $\checkmark$ & & \\
NeO 360 \cite{irshadNeO360Neural2023} & & & & & & & & & & \\
6Img-to-3D \cite{gieruc6Imgto3DFewImageLargeScale2025} & & & & & & & & & & \\
\midrule
\multicolumn{11}{l}{\textbf{3D Gaussian Splatting (3DGS) Approaches}} \\
\midrule
3DGS \cite{kerbl2023threedgs} & & $\checkmark$ & & & & & & & & \\
FSGS \cite{zhuFSGSRealTimeFewshot2024} & $\checkmark$ & $\checkmark$ & & & & & & & & \\
CoR-GS \cite{zhangCoRGSSparseView3D2024} & $\checkmark$ & $\checkmark$ & & & & & & & & \\
DNGaussian \cite{liDNGaussianOptimizingSparseView2024} & & & & & & & & & & \\
Speedy-Splat \cite{hansonSpeedySplatFast3D} & & & $\checkmark$ & & & & & & & \\
Point Cloud Densification \cite{chanPointCloudDensification2024} & & $\checkmark$ & & & & $\checkmark$ & & & & \\
Improving Geometry... \cite{kimImprovingGeometrySparseView2024} & & $\checkmark$ & $\checkmark$ & & & & $\checkmark$ & & & \\
LoopSparseGS \cite{baoLoopSparseGSLoopBased2024} & & $\checkmark$ & & & & & & & & \\
Optimizing 3DGS... \cite{chenOptimizing3DGaussian2024} & & & & & & & & & & \\
PointGS \cite{xiangPointGSPointAttentionAware2025} & & & & & & & & & & \\
JointSplat \cite{xiaoJointSplatProbabilisticJoint2025} & & & & & $\checkmark$ & & & & & $\checkmark$ \\
SPARS3R \cite{tangSPARS3RSemanticPrior} & & & & & & & & & & \\
VGNC \cite{linVGNCReducingOverfitting2025} & $\checkmark$ & $\checkmark$ & $\checkmark$ & & & & & & & \\
Intern-GS \cite{sunInternGSVisionModel2025} & $\checkmark$ & $\checkmark$ & $\checkmark$ & & & & & & & \\
TranSplat \cite{zhangTranSplatGeneralizable3D} & $\checkmark$ & & & & $\checkmark$ & & & & & $\checkmark$ \\
SparSplat \cite{jenaSparSplatFastMultiView2025} & $\checkmark$ & & $\checkmark$ & & & & & & & \\
HiSplat \cite{tangHiSplatHierarchical3D2024} & & & & & $\checkmark$ & & & & & \\
Sparse2DGS \cite{wuSparse2DGSGeometryPrioritizedGaussian} & $\checkmark$ & & & & & & & & & \\
UniForward \cite{tianUniForwardUnified3D2025} & & & & & & $\checkmark$ & & & & \\
Deceptive-NeRF/3DGS \cite{liuDeceptiveNeRF3DGSDiffusionGenerated2024} & & & & & & & & & & \\
LM-Gaussian \cite{yuLMGaussianBoostSparseview2024} & & & & & & & & & & \\
FlowR \cite{fischerFlowRFlowingSparse2025} & & & & & $\checkmark$ & $\checkmark$ & & & & \\
AugGS \cite{duAugGSSelfaugmentedGaussians2024} & & & & & & & & & & \\
V3D \cite{chenV3DVideoDiffusion2024} & & & & & & & & & & \\
CAT3D \cite{gaoCAT3DCreateAnything2024} & & & & & & & & & & \\
CAT4D \cite{wuCAT4DCreateAnything2024} & & & & & & & & & & \\
ProSplat \cite{luProSplatImprovedFeedForward2025} & & & & & $\checkmark$ & & & & & \\
InstantSplat \cite{fanInstantSplatUnboundedSparseview} & & & $\checkmark$ & & & & $\checkmark$ & & & \\
CF-3DGS \cite{fuCOLMAPFree3DGaussian} & & & & $\checkmark$ & & & & & & \\
MV-DUSt3R+ \cite{tangMVDUSt3RSingleStageScene} & & & & & & & & & & \\
Seeing A 3D World... \cite{zhangSeeing3DWorld2025} & & & & & & & & & & \\
FreeSplatter \cite{xuFreeSplatterPosefreeGaussian2024} & & & & & & & & & & \\
Gaussian Scenes \cite{Paul2024GaussianSP} & & & & & & & & & & \\
GaussianObject \cite{yangGaussianObjectHighQuality3D2024} & & $\checkmark$ & & & & & & & $\checkmark$ & \\
Free360 \cite{zhangFree360LayeredGaussian} & & $\checkmark$ & $\checkmark$ & & & & & & & \\
SpatialSplat \cite{shengSpatialSplatEfficientSemantic2025} & & & & & & & & & & \\
UniGS \cite{Wu2024UniGS} & & & & & & & & & & \\
\midrule
\multicolumn{11}{l}{\textbf{Diffusion \& VFM Integration}} \\
\midrule
Sparse3D \cite{zouSparse3DDistillingMultiviewConsistent2024} & & & & $\checkmark$ & & & & & & \\
GenFusion \cite{wuGenFusionClosingLoop2025} & & $\checkmark$ & & & & & & & & \\
VI3DRM \cite{chenVI3DRMMeticulous3D2024} & & & & & & & & & $\checkmark$ & \\
Fine-tuning Diffusion... \cite{tangFinetuningDiffusionModel2024} & & & & & & & & & & \\
Generating Material-Aware... \cite{maoGeneratingMaterialAware3D} & & & & & & & & & & \\
Sp2360 \cite{paulSp2360Sparseview3602024} & & $\checkmark$ & & & & & & & & \\
SIR-DIFF \cite{maoSIRDIFFSparseImage} & & & & & & & & & & \\
iFusion \cite{wuIFusionInvertingDiffusion2023} & & & & & & & & & & \\
How to Use Diffusion Priors... \cite{Wang2024HowTU} & $\checkmark$ & $\checkmark$ & & & & & & & & \\
SparseFusion \cite{zhouSparseFusionDistillingViewConditioned2023} & & & & $\checkmark$ & & & & & & \\
\bottomrule
\end{tabular}%
}
\caption{Dataset Coverage of Sparse-View 3D Reconstruction Methods. A '$\checkmark$' indicates that the method was evaluated on the respective benchmark dataset according to its primary publication.}
\label{tab:dataset_coverage}
\end{table*}

\section{Comparative Evaluation (Datasets and Metrics)}
\label{sec:comparative}

Rapid progress in sparse-view 3D reconstruction calls for standardized evaluation protocols to enable objective comparisons of methods. This section outlines the most commonly used benchmark datasets and quantitative metrics and provides a synthesized overview of the state-of-the-art performance of these benchmarks.

\subsection{Benchmark Datasets}
\label{ssec:datasets}

Evaluating sparse-view 3D reconstruction methods requires a range of datasets, each introducing different challenges related to scene complexity, object diversity, camera setup, and lighting conditions.

\begin{itemize}
    \item \textbf{DTU (Technical University of Denmark MVS Dataset)}: A controlled laboratory dataset with precise ground-truth 3D models and camera poses~\cite{jensen2014dtu}. It is commonly used to assess reconstruction quality and generalization, especially in sparse-view settings (e.g., three views)\cite{zhuFSGSRealTimeFewshot2024,zhangCoRGSSparseView3D2024}.
    \item \textbf{LLFF (Light Field-based Forward-Facing Dataset)} / \textbf{Mip-NeRF 360}: Real-world scenes captured with inward-facing, 360-degree images\cite{mildenhall2019llff}. These datasets present challenges owing to large scene bounds, depth variation, and occlusion~\cite{yu2021pixelnerf,baoLoopSparseGSLoopBased2024}.
    \item \textbf{Tanks and Temples}: Large-scale outdoor scenes with complex geometry~\cite{knapitsch2017tanks}, captured with professional equipment. It is widely used to evaluate robustness and scalability in sparse-view scenarios~\cite{fanInstantSplatUnboundedSparseview,fuCOLMAPFree3DGaussian}.
    \item \textbf{CO3D}: A large-scale collection of everyday objects from diverse viewpoints~\cite{reizenstein2021co3d}, supporting generalizable 3D reconstruction and category-level evaluation in sparse-view conditions~\cite{zouSparse3DDistillingMultiviewConsistent2024,fuCOLMAPFree3DGaussian}.
    \item \textbf{RealEstate10K} / \textbf{DL3DV-10K}: Datasets for wide-baseline and 360-degree novel view synthesis in real-world scenes~\cite{zhou2018stereomag,ling2023dl3dv10k}, featuring diverse camera motions and layouts. DL3DV-10K is a recent benchmark for state-of-the-art sparse-view studies~\cite{chenMVSplat360FeedForward3602024,luProSplatImprovedFeedForward2025}.
    \item \textbf{ScanNet / ScanNet++}: Indoor scene datasets with cluttered environments and complex layouts~\cite{dai2017scannet,tang2024scannetpp}. It is widely used to test reconstruction under challenging indoor conditions~\cite{tanNOPESACNeuralOnePlane2023,tangMVDUSt3RSingleStageScene}.
    \item \textbf{MVImgNet}: A diverse benchmark of multi-view images spanning a wide range of objects and scenes~\cite{yu2023mvimgnet}, used to test generalization across domains~\cite{fanInstantSplatUnboundedSparseview}.
    \item \textbf{Other Specialized Datasets}: Includes synthetic datasets like ShapeNet~\cite{chang2015shapenet} for controlled shape consistency and generalization, OmniObject3D~\cite{wu2023omniobject3d} and OpenIllumination~\cite{Liu2023OpenIlluminationAM} for relightable or view-consistent reconstruction, and ACID~\cite{xiao2021acid} for real-world, instance-level 3D learning. Additional datasets address niche scenarios, such as miniature scene reconstruction and dynamic urban environments, for the targeted evaluation of challenging domains.
\end{itemize}
The coverage of these benchmark datasets by various sparse-view 3D reconstruction methods is comprehensively summarized in Table \ref{tab:dataset_coverage}.

\subsection{Quantitative Metrics}
\label{ssec:metrics}

Performance is typically evaluated using a combination of metrics covering visual quality, geometric accuracy, and computational efficiency.

\begin{itemize}
\item \textbf{Rendering Quality Metrics}:
    \begin{itemize}
    \item \textbf{PSNR (Peak Signal-to-Noise Ratio)}: Measures image reconstruction fidelity; higher is better\cite{huynh2008psnr}.
    \item \textbf{SSIM (Structural Similarity Index Measure)}: Assesses perceptual similarity between images; higher values indicate greater similarity\cite{wang2004ssim}.
    \item \textbf{LPIPS (Learned Perceptual Image Patch Similarity)}: Computes perceptual distance using deep features; lower values indicate higher perceptual similarity\cite{zhang2018lpips}.
    \item \textbf{FID (Frechet Inception Distance)}: Evaluates similarity between distributions of generated and real images; lower is better\cite{heusel2017fid}.
    \item \textbf{DISTS}: A differentiable perceptual similarity metric\cite{ding2020dists}.
\end{itemize}
\item \textbf{Geometric Accuracy Metrics}:
    \begin{itemize}
    \item \textbf{Chamfer Distance (CD)}: Compares similarity between point clouds or shapes; lower is better\cite{fan2017chamfer}.
    \item \textbf{Normal Consistency (NC) / F-Score (FS)}: Assess surface and normal reconstruction quality\cite{mescheder2019normal}.
    \item \textbf{RPEt (Relative Pose Error - translation)}: Measures translational camera pose error; lower values are better\cite{sturm2012benchmark}.
    \item \textbf{RPEr (Relative Pose Error - rotation)}: Measures rotational camera pose error; lower values are better\cite{engel2014lsdslam}.
    \item \textbf{ATE (Absolute Trajectory Error)}: RMSE between estimated and ground-truth camera trajectories; lower is better\cite{sturm2012benchmark}.
    \item \textbf{PDC (Patch-wise Depth Pearson Correlation)}: Correlates rendered and estimated depth maps to assess local geometric plausibility\cite{yuan2022patch}.
\end{itemize}
\item \textbf{Efficiency Metrics}:
\begin{itemize}
    \item \textbf{Training Time / Inference Time}: Quantifies computational cost; lower is better.
    \item \textbf{FPS (Frames Per Second)}: Measures rendering speed; higher is better.
    \item \textbf{Number of Gaussians (GS Num)}: Indicates scene representation density in 3DGS; lower values suggest greater efficiency.
\end{itemize}
\end{itemize}

\begin{table*}[p]
\centering
\scriptsize
\renewcommand{\arraystretch}{1.3}
\rowcolors{2}{rowgray}{white}
\label{tab:in_depth_analysis}
\resizebox{\textwidth}{!}{%
\begin{tabular}{p{2.5cm}|p{4.5cm}|p{5.5cm}|p{4.5cm}}
\toprule
\textbf{Method (Year)} & \textbf{Core Innovation / Key Mechanism} & \textbf{Sparse-View Contribution / Strengths} & \textbf{Limitations} \\
\midrule
\multicolumn{4}{l}{\textbf{\textit{Geometry-Based Methods}}} \\
\midrule
\textbf{NOPE-SAC (2023)} \cite{tanNOPESACNeuralOnePlane2023} & Neural One-Plane RANSAC learns pose hypotheses from minimal 3D plane correspondences. & Addresses insufficient correspondences for robust pose estimation in sparse 2-view settings (e.g., indoor scenes, low texture). It improves the camera pose and planar reconstruction accuracy. & Primarily for planar scenes. It may struggle with highly nonplanar structures. \\
\addlinespace
\textbf{SparseCraft (2024)} \cite{younesSparseCraftFewShotNeural2024} & Stereopsis-guided geometric linearization regularizes implicit SDF learning using MVS-derived normals and colors. & Achieves high-fidelity few-shot reconstruction and NVS rapidly (e.g., 9 min of training). Robust noise from MVS cues bridging photogrammetry and deep learning. & Relies on MVS cues, inheriting MVS limitations (e.g., sufficient overlap and non-Lambertian surfaces). \\
\addlinespace
\textbf{3DFIRES (2024)} \cite{jin3DFIRESFewImage2024} & Fuses multi-view information at feature level using a Directed Ray Distance Function (DRDF) with a Vision Transformer backbone. & Reconstructs complete 3D geometry, including hidden surfaces, from as few as one posed image. It generalizes well to varying view counts and unseen scenes. & Potential for artifacts if feature fusion is imperfect. Computational cost of transformer backbone. \\
\midrule
\multicolumn{4}{l}{\textbf{\textit{Neural Implicit Representations (NeRF and Variants)}}} \\
\midrule
\textbf{RegNeRF (2022)} \cite{niemeyerRegNeRFRegularizingNeural2022} & Patch-based geometry regularization (smoothness loss on depth) and appearance regularization via normalizing flow model. Sample-space annealing strategy. & Mitigates overfitting and "floating artifacts" in sparse-view NeRF. It enhances geometric consistency and color prediction from limited inputs. & Can be slow owing to per-scene optimization. It assumes fixed camera poses. \\
\addlinespace
\textbf{ZeroRF (2023)} \cite{shiZeroRFFastSparse2023} & Integrates a tailored Deep Image Prior into a factorized NeRF representation. & Achieves fast (seconds to minutes), high-quality 360$^\circ$ reconstruction from very few views (4-6). This eliminates the need for pretraining or explicit regularization. & Primarily designed for 360$^\circ$ scenes. It may not generalize to other scene types as effectively. \\
\addlinespace
\textbf{pixelNeRF (2021)} \cite{yu2021pixelnerf} & Conditions a NeRF representation on input images in a fully convolutional manner, learning a scene prior across multiple scenes. & Enables generalizable feed-forward novel view synthesis from one or few images. It eliminates lengthy per-scene optimization. & may not achieve the absolute highest fidelity of per-scene optimized NeRFs. The performance can vary significantly depending on the input view quality. \\
\midrule
\multicolumn{4}{l}{\textbf{\textit{3D Gaussian Splatting (3DGS) Approaches}}} \\
\midrule
\textbf{InstantSplat (2024)} \cite{fanInstantSplatUnboundedSparseview} & Integrates dense stereo models (DUSt3R) for coarse initialization with a fast 3D-Gaussian optimization (F-3DGO) module. & Achieves rapid (<1 min), high-quality, pose-free 3DGS reconstruction for unbounded scenes from sparse input. Superior rendering quality and pose estimation accuracy. & May miss extremely fine details compared with dense methods. \\
\addlinespace
\textbf{SparseGS (2025)} \cite{xiongSparseGSRealTime360deg2025} & Novel depth rendering techniques, patch-based depth correlation loss, Unseen Viewpoint Regularization (UVR) via SDS, and advanced floater pruning. & Addresses "floaters" and "background collapse" in sparse-view 3DGS. Achieves SOTA performance in 360$^\circ$ and forward-facing sparse view synthesis (3-12 views). & Relies on effective depth priors and SDS guidance, which can be sensitive to the hyperparameter tuning. \\
\addlinespace
\textbf{CoR-GS (2024)} \cite{zhangCoRGSSparseView3D2024} & Novel "co-regularization" perspective: simultaneously trains two 3DGS fields, leveraging their point and rendering disagreement for self-supervision. & Combats overfitting in sparse-view 3DGS. It regularizes scene geometry, reconstructs compact representations, and achieves SOTA NVS quality while reducing the Gaussian count. & Requires the simultaneous training of two models, potentially increasing the memory footprint during training. \\
\midrule
\multicolumn{4}{l}{\textbf{\textit{Diffusion and VFM Integration}}} \\
\midrule
\textbf{Sparse3D (2024)} \cite{zouSparse3DDistillingMultiviewConsistent2024} & Distills robust priors from a multiview-consistent diffusion model (guided by epipolar controller) to refine a neural radiance field. & Delivers high-quality, perceptually sharp results for object reconstruction from extremely sparse views ( 2-3 images). It shows strong generalization to unseen categories & Susceptibility to the "Janus problem" and struggles with extreme partial observations or thin structures. Relies on accurate camera poses. \\
\addlinespace
\textbf{GenFusion (2025)} \cite{wuGenFusionClosingLoop2025} & Reconstruction-driven video diffusion model learns to condition on RGB-D renderings in a cyclical fusion pipeline. & Bridges 3D reconstruction and generation. Restoration frames are iteratively added to densify sparse input and address viewpoint saturation. The effectiveness of the proposed method for sparse-view synthesis was validated. & Denoising steps increase time. Potential for blurriness in large invisible regions if the generative model fails. \\
\addlinespace
\textbf{VI3DRM (2024)} \cite{chenVI3DRMMeticulous3D2024} & Diffusion-based model operating within an ID-consistent and perspective-disentangled 3D latent space. & Generates exceptionally realistic and photorealistic novel views and constructs accurate point maps/meshes. It disentangles semantic, color, material, and lighting information. & Performance may be sensitive to the quality of the learned latent spaces. \\
\addlinespace
\textbf{Sp2360 (2024)} \cite{paulSp2360Sparseview3602024} & Uses cascaded 2D diffusion priors in an iterative process to augment sparse views for 360$^\circ$ scene reconstruction. & Efficiently reconstructs 360$^\circ$ scenes from very limited views (e.g., 9 inputs). Diffusion models perform inpainting and artifact elimination to ensure multiview consistency. & Relies on the quality of the 2D diffusion model and its ability to maintain 3D consistency across views. \\
\bottomrule
\end{tabular}%
}
\caption{In-Depth Analysis of Representative Sparse-View 3D Reconstruction Methods. This table highlights the core innovations, specific contributions to sparse-view challenges, and identified limitations of key methods across different paradigms.}
\label{tab:in_depth_analysis_improved}
\end{table*}

\subsection{Comparative Analysis}
\label{ssec:comparative-analysis}

The performance landscape of sparse-view 3D reconstruction is evolving rapidly, with leading methods demonstrating distinct strengths across a range of benchmarks. Figure~\ref{fig:radar_chart} provides a visual comparison of these methods based on the normalized evaluation metrics, offering a clear perspective on their relative performances. Table~\ref{tab:in_depth_analysis_improved} presents an in-depth analysis of representative approaches, detailing core innovations, input sparsity, pose requirements, representation types, and runtime characteristics. The table further identifies the main contributions of each method to sparse-view reconstruction, outlines the key trade-offs, and summarizes the remaining limitations of different paradigms.

\begin{figure*}[t]  % use [t] or [b] for top or bottom placement
    \centering
    \includegraphics[width=\textwidth]{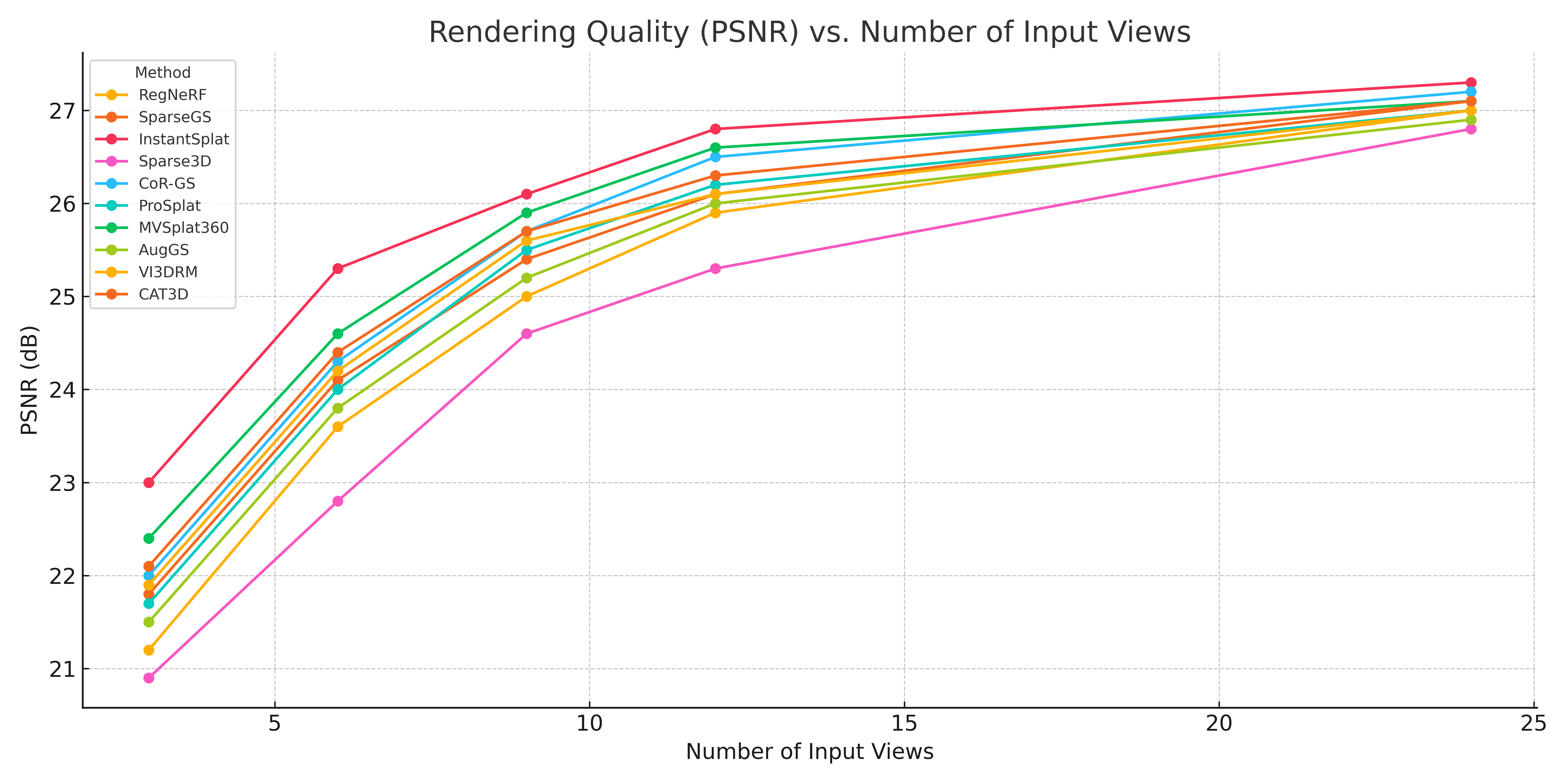}
    \caption{Rendering quality (PSNR) versus number of input views for leading sparse-view 3D reconstruction methods. The plot compares NeRF-based, 3DGS-based, and hybrid approaches under varying input sparsity, highlighting each method's robustness and data efficiency.}
    \label{fig:psnr_number_of_views}
\end{figure*}

Increasing the number of input views consistently improves both visual quality, measured by higher PSNR and SSIM and lower LPIPS, and geometric accuracy, reflected in lower pose errors and Chamfer Distance, across most methods. The most substantial gains were observed when increasing from a very sparse input (such as 3 views) to a moderately sparse regime (6–9 views). Figure~\ref{fig:psnr_number_of_views} shows how the rendering quality (PSNR) increases as the number of input views increases for several leading sparse-view 3D reconstruction methods.

\section{Current Challenges}
\label{sec:challenges}

Despite significant progress, sparse-view 3D reconstruction continues to face several challenges. These unresolved problems remain at the forefront of research, driving the development of novel algorithms and hybrid approaches. Although recent advances have improved both quality and efficiency, the fundamental limitations imposed by sparse inputs create enduring obstacles that current methods have yet to overcome.

\subsection{Fundamental Ambiguities and Artifacts}
\label{sssec:ambiguities-artifacts}

Sparse input views create ill-posed reconstruction problems, allowing multiple plausible 3D solutions for the same set of 2D observations \cite{fanInstantSplatUnboundedSparseview}. This ambiguity produces common artifacts, such as "floaters" (spurious geometry in empty space), blurry textures, and "background collapse," where distant backgrounds are incorrectly projected closer to the camera \cite{xiongSparseGSRealTime360deg2025, gieruc6Imgto3DFewImageLargeScale2025}. The "Janus problem" an inconsistent or multi-faced hallucination of unseen surfaces, remains a major limitation for generative models \cite{zouSparse3DDistillingMultiviewConsistent2024}. Reconstructing occluded or hidden surfaces from minimal inputs is especially difficult because there is often insufficient data to reliably infer the missing geometry \cite{jin3DFIRESFewImage2024}.

\subsection{Computational Bottlenecks and Scalability}
\label{sssec:computational-scalability}

Although explicit methods such as 3DGS have enabled real-time rendering, achieving real-time performance for the entire reconstruction process, including training and optimization, remains a major challenge, particularly for large-scale or high-resolution scenes \cite{luProSplatImprovedFeedForward2025}. Many state-of-the-art methods require significant computation and long training times, limiting their practicality for real-world or dynamic applications \cite{dengDepthsupervisedNeRFFewer2022, hamdiSPARFLargeScaleLearning2023}. In addition, the large memory requirements for storing volumetric or dense point cloud representations can restrict scalability, particularly when working with complex or extensive scenes \cite{wangDiffusionModels3D2025}.

\subsection{Robustness to ``In-the-Wild'' Conditions}
\label{sssec:in-the-wild-robustness}

Real-world data rarely match the controlled laboratory benchmark conditions. Sparse-view inputs often include uncalibrated images, noisy or imprecise camera poses, and dynamic or moving scene elements, all of which undermine traditional SfM pipelines \cite{fanInstantSplatUnboundedSparseview, fuCOLMAPFree3DGaussian, weiSemanticallyAwareMultiView2024}. Scenes with challenging materials, such as specular, transparent, and low-textured surfaces, are particularly problematic. These surface types hinder feature matching and disrupt photometric consistency \cite{maoGeneratingMaterialAware3D, zhangSeeing3DWorld2025, kimImprovingGeometrySparseView2024}. Real-world capture also introduces noise, outliers, and varying illumination, which further degrade the reconstruction quality. Addressing these uncontrolled conditions requires models that demonstrate much greater robustness than the current solutions typically provide \cite{gieruc6Imgto3DFewImageLargeScale2025}.

\subsection{Generalization Across Diverse Scenes and Objects}
\label{sssec:generalization-diversity}

Many leading methods remain dependent on per-scene optimization. They require retraining or fine-tuning for every new scene, which is both time-consuming and impractical for open-world deployment. Although some generalizable approaches exist, they often struggle to maintain high fidelity across different object categories, scene types (such as indoor vs. outdoor and object-centric vs. unbounded), and extreme changes in viewpoint. This can result in blurry or perceptually inconsistent reconstructions, particularly for out-of-distribution data \cite{hamdiSPARFLargeScaleLearning2023, zouSparse3DDistillingMultiviewConsistent2024}. Building models that generalize well from limited examples is crucial for scaling sparse-view 3D reconstruction to real-world applications.

\section{Promising Emerging Trends}
\label{ssec:emerging-trends}

The field of sparse-view 3D reconstruction is evolving rapidly, driven by interdisciplinary techniques and fundamental methodological advances.

\subsection{Leveraging Advanced Generative Priors}
\label{sssec:generative-priors}

A major trend is the growing use of powerful generative models, particularly diffusion models, to synthesize missing content and provide strong priors for reconstructions \cite{wangDiffusionModels3D2025}. These models, pre-trained on large-scale datasets, can create “3D-aware images” or “pseudo-observations” that densify sparse inputs, resulting in higher-quality and more complete reconstructions \cite{tangFinetuningDiffusionModel2024, liuDeceptiveNeRF3DGSDiffusionGenerated2024}. State-of-the-art methods such as \textbf{Sp2360} \cite{paulSp2360Sparseview3602024} and \textbf{GenFusion} \cite{wuGenFusionClosingLoop2025} demonstrate cascaded or cyclical refinement of 3D scenes using 2D or video diffusion models, achieving improved multi-view consistency. \textbf{VI3DRM} \cite{chenVI3DRMMeticulous3D2024} highlights the potential of diffusion models to learn perspective-disentangled latent spaces, enabling highly realistic novel view synthesis. Additionally, the integration of Vision Foundation Models (VFMs), such as CLIP, SAM, and DINO, provides rich semantic and feature-level guidance, supporting robust model initialization and optimization \cite{radford2021clip, kirillov2023sam, caron2021dino}.

\subsection{Hybrid Representations and Architectures}
\label{sssec:hybrid-architectures}

There has been a clear shift toward hybrid approaches that combine multiple types of 3D representations. Integrating explicit structures (such as point clouds or Gaussians) with implicit neural fields (such as NeRFs or SDFs) enables methods to overcome individual weaknesses and exploit their complementary strengths. For example, \textbf{SparseCraft} \cite{younesSparseCraftFewShotNeural2024} uses Multi-View Stereo (MVS) cues to regularize implicit SDF-based models, while \textbf{X-NeRF} \cite{zhuXNeRFExplicitNeural2022} and \textbf{6Img-to-3D} \cite{gieruc6Imgto3DFewImageLargeScale2025} incorporate explicit scene completion or triplane representations for more generalizable rendering. These hybrid strategies yield improvements in speed, memory efficiency, and geometric accuracy, and are increasingly prominent in state-of-the-art research.

\subsection{Joint Learning of Geometry and Pose}
\label{sssec:joint-learning}

A major direction is the joint optimization of geometry and camera poses, which addresses the classic "chicken-and-egg" dilemma of sparse-view reconstruction. Recent methods, such as \textbf{InstantSplat} \cite{fanInstantSplatUnboundedSparseview}, \textbf{CF-3DGS} \cite{fuCOLMAPFree3DGaussian}, and \textbf{FreeSplatter} \cite{xuFreeSplatterPosefreeGaussian2024}, optimize 3D scene structure and camera parameters together. These “pose-free” or “COLMAP-free” frameworks increase robustness to poor initializations and reduce dependence on traditional, slow SfM pipelines. End-to-end learning of pose and geometry enables fast, accurate reconstruction from uncalibrated images, as also shown by \textbf{MV-DUSt3R+} \cite{tangMVDUSt3RSingleStageScene} and \textbf{SparseAGS} \cite{zhaoSparseviewPoseEstimation2024}.

\subsection{Adaptive and Optimized Training Strategies}
\label{sssec:adaptive-strategies}

Modern pipelines go beyond fixed architectures and adopt training strategies tailored to sparse data. These include sample space annealing and regularization schedules, as in \textbf{RegNeRF} \cite{niemeyerRegNeRFRegularizingNeural2022}, and dynamic control over representation size, exemplified by \textbf{VGNC}'s validation-guided Gaussian number control \cite{linVGNCReducingOverfitting2025}. Many methods now use iterative refinement, self-correction, and error detection loops during training \cite{wuGenFusionClosingLoop2025}. These advances make optimization more stable and the results more reliable, even with minimal inputs.

\subsection{Semantic and Material-Aware Reconstruction}
\label{sssec:semantic-material-aware}

Another growing trend is the enrichment of reconstructions with high-level semantic and material properties. For example, \textbf{Wei et al.}’s\cite{weiSemanticallyAwareMultiView2024} semantically aware multiview pipeline improves dense reconstruction by enforcing semantic consistency, which is especially valuable for dynamic real-world environments. Similarly, \textbf{Mao et al.}  \cite{maoGeneratingMaterialAware3D} disentangled geometry, material, and lighting for material-aware 3D asset creation, enabling relightable and physically meaningful outputs. These advances allow for more informative 3D reconstructions and enable downstream applications such as semantic editing, object recognition, and photorealistic relighting.

\subsection{Real-time Performance and Efficiency Optimization}
\label{sssec:real-time-efficiency}

Practical 3D reconstruction requires high speed and efficiency. Methods such as \textbf{Speedy-Splat} \cite{hansonSpeedySplatFast3D} optimize the rendering operations and prune unnecessary Gaussians, significantly accelerating 3DGS pipelines. \textbf{DNGaussian} \cite{liDNGaussianOptimizingSparseView2024} and \textbf{FSGS} \cite{zhuFSGSRealTimeFewshot2024} achieve fast training and inference through advanced regularization and efficient Gaussian management. Feed-forward models such as \textbf{SparSplat} \cite{jenaSparSplatFastMultiView2025} and \textbf{UniForward} \cite{tianUniForwardUnified3D2025} deliver real-time, generalizable performance from sparse and even uncalibrated inputs. These advances mark major progress toward instant 3D reconstruction and its real-world applications.

In summary, sparse-view 3D reconstruction faces persistent challenges, ranging from ill-posed ambiguities to computational demands and generalization problems. However, this field is rapidly advancing. Innovations in generative modeling, hybrid representations, and adaptive optimization converge to provide robust, efficient, and semantically meaningful solutions. This progress paves the way for widespread 3D content creation and intelligent scene understanding in diverse, real-world scenarios.

\section{Future Research Directions}
\label{sec:future}

Despite rapid progress, sparse-view 3D reconstruction faces several fundamental challenges. This section builds on the issues and trends discussed in Section~\ref{sec:challenges} and proposes creative and impactful directions for future research. The goal is to bridge the gap between research advances and real-world deployment. Ultimately, the aim is to achieve a high-fidelity 3D understanding from minimal observational data, making robust 3D reconstruction accessible and practical for diverse applications.

\subsection{Unified Multi-Modal Generative 3D Priors}
\label{ssec:unified-generative-priors}

Recent methods often use 2D diffusion models to fill in missing views through score distillation or pseudo-observation generation. However, these approaches are inherently limited by their reliance on a 2D image space. They often struggle to guarantee true 3D consistency and geometric accuracy.

\begin{itemize}
\item \textbf{3D-Native Generative Foundation Models}: Future work should focus on generative models trained directly on large-scale, diverse 3D datasets. These “3D-native” models would learn geometry, topology, and physical properties, enabling high-fidelity scene inference from very sparse data, such as a single image or even text. Unlike current methods, they can synthesize a full 3D structure with intrinsic consistency, not just 2D projections \cite{wangDiffusionModels3D2025}.
\item \textbf{Integrated Multi-Modal Generative Priors}: Next-generation generative models should produce more than just RGB images. By learning to generate RGB-D, normal maps, semantic masks, and material attributes in a unified and 3D-consistent manner, these models can provide richer priors. This would better constrain downstream 3D reconstruction and boost both geometric accuracy and semantic understanding.
\item \textbf{Disentangled 3D Latent Spaces}: Research into disentangled latent representations such as those explored in \texttt{VI3DRM} \cite{chenVI3DRMMeticulous3D2024} should continue. Highly disentangled spaces would enable independent control over identity, geometry, texture, material, and illumination. This flexibility enables robust 3D content generation, editing, and manipulation from sparse inputs.
\end{itemize}

\subsection{Robustness and Fidelity in Extreme "In-the-Wild" Conditions}
\label{ssec:extreme-in-the-wild-robustness}

Significant hurdles remain in sparse-view 3D reconstruction in unconstrained, dynamic, and real-world environments.
\begin{itemize}
\item \textbf{Dynamic Scene Reconstruction from Unstructured Streams}: Reconstructing deforming objects and dynamic scenes from sparse, uncalibrated, and unsynchronized video streams (e.g., multiple handheld phone captures) is largely unsolved. Future work should enable the joint estimation of 4D geometry, motion, and camera trajectories, moving beyond the static scene assumptions.
\item \textbf{Illumination and Material-Agnostic Inverse Rendering}: Disentangling geometry, material properties (including complex BRDFs like translucency and specularity), and environmental illumination from sparse, real-world images remains extremely challenging~\cite{maoGeneratingMaterialAware3D}. More robust neural inverse rendering techniques are required to accurately infer these properties under limited, uncalibrated, and variable lighting conditions.
\item \textbf{Noise-Robust and Degraded Data Reconstruction}: Developing methods resilient to sensor imperfections such as motion blur, atmospheric effects, lens distortions, variable noise, and low dynamic range remains an open challenge. Progress requires learning robust feature representations and reconstruction priors that can handle significant data degradation and move beyond the idealized capture conditions.
\end{itemize}

\subsection{Towards Real-time, On-Device, and Continual 3D Understanding}
\label{ssec:real-time-continual}

The goal is to achieve a ubiquitous, instantaneous, and persistent 3D understanding in practical settings.
\begin{itemize}
\item \textbf{Ultra-Efficient On-Device Pipelines}: Optimize the entire sparse-view 3D reconstruction process for resource-constrained edge devices such as smartphones and AR/VR headsets. This requires lightweight neural architectures, sparse data structures, hardware-aware designs, and efficient optimization, as demonstrated in works such as \texttt{Speedy-Splat}\cite{hansonSpeedySplatFast3D} and \texttt{DNGaussian}\cite{liDNGaussianOptimizingSparseView2024}.
\item \textbf{Adaptive Level-of-Detail (LoD) and Streaming Reconstruction}: For large-scale environments, develop adaptive LoD mechanisms that stream and reconstruct 3D content at varying resolutions depending on viewpoint, computational budget, and network bandwidth. Methods should support seamless transitions between LoDs and efficient data management for scalable and real-world deployment.
\item \textbf{Continual Learning and Living 3D Maps}: Enable "living" 3D maps that are updated continuously as new sparse observations become available. This requires robust change detection, incremental reconstruction, efficient data association, and consistency maintenance in dynamic, long-term scenarios. This capability is critical for applications such as autonomous navigation and the development of digital twins.
\end{itemize}

\subsection{Intelligent Acquisition and Human-in-the-Loop Reconstruction}
\label{ssec:intelligent-acquisition}

Future systems can move beyond passive reconstruction to include active and interactive processes.
\begin{itemize}
\item \textbf{Uncertainty-Aware Active Reconstruction}: Develop models that explicitly quantify uncertainty in their outputs and use this information to guide adaptive view acquisition. Intelligent systems, such as drones and robotic agents, can target unobserved or ambiguous regions to reduce uncertainty, thereby enabling more efficient and complete reconstructions from minimal inputs.
\item \textbf{Human-in-the-Loop Refinement and Editing}: Integrate intuitive user interfaces that allow users to interactively guide the reconstruction, correct errors, or enhance details in difficult regions. This "human-in-the-loop" paradigm combines automated methods with human expertise, improving fidelity and enabling creative control, including real-time semantic editing and relighting of 3D scenes~\cite{tianUniForwardUnified3D2025}.
\end{itemize}

These directions represent a shift towards intelligent, adaptive, and user-centric 3D reconstruction systems, unlocking new capabilities for content creation, immersive experiences, and real-world autonomous applications.

\section{Conclusion and Discussion}
\label{sec:conclusion}

Sparse-view 3D reconstruction remains one of the most fundamental and challenging problems in computer vision, requiring innovative solutions to recover detailed 3D geometries and photorealistic appearances from limited and ambiguous 2D observations. This survey traces the evolution of the field, covering early geometry-based methods, the emergence of neural implicit representations (NeRFs), and the latest advances in 3D Gaussian Splatting (3DGS).

Each methodological family targets the key obstacles posed by sparse data occlusion, pose uncertainty, overfitting, and limited supervision. Although traditional SfM and MVS methods are foundational, they have critical limitations in low-overlap scenarios. The rise of NeRFs and their variants introduced implicit volumetric modeling and regularization techniques that addressed sparsity by leveraging geometric and learned priors. The 3DGS has rapidly become a state-of-the-art paradigm that combines high efficiency and real-time rendering with robust solutions for sparse-view settings.

A transformative trend across all paradigms is the integration of generative diffusion models and powerful Vision Foundation Models (VFMs). These approaches provide robust priors, synthesize plausible geometries and textures in unseen regions, and enable the creation of high-quality pseudo-observations that densify the sparse data. Hybrid strategies, joint optimization of pose and geometry, and efficient pipelines have further advanced the applicability and performance of sparse-view 3D reconstruction.

The field is undergoing a significant transition from dependence on dense, well-calibrated inputs and heavy per-scene optimization to flexible, generalizable, and efficient methods that can operate with minimal data and fewer constraints. However, notable challenges remain, particularly in achieving robust generalization across diverse domains, handling complex real-world conditions, and integrating deep semantic understanding into 3D modeling.

This survey aims to serve as a comprehensive reference for researchers and practitioners. By analyzing state-of-the-art methods, identifying unresolved challenges, and highlighting emerging trends such as 3D-native generative models, intelligent acquisition, and continual learning, we aim to inspire ongoing innovation. Our goal is to accelerate the development and deployment of robust, intelligent 3D reconstruction systems that enable rich, actionable 3D representations in everyday applications

\bibliographystyle{IEEEtran}
\bibliography{refrences} 

\end{document}